\documentclass{article}

\usepackage[
    colorlinks=true,
    citecolor=blue,
    linkcolor=blue,
    urlcolor=black
]{hyperref}
\usepackage[preprint]{neurips_2026}
\usepackage{hyperref}
\usepackage{algorithm}
\usepackage{algpseudocode}
\usepackage{xcolor}
\usepackage{listings}
\usepackage{caption}
\usepackage{makecell}
\usepackage{wrapfig}
\usepackage{booktabs}
\usepackage{float}
\usepackage[table]{xcolor}
\usepackage{longtable}

\usepackage{amsmath}
\usepackage{booktabs}
\usepackage{graphicx}
\usepackage{cleveref}
\newcommand{\std}[1]{\textnormal{\scriptsize $\pm$ #1}}
\newcommand{\twodigit}[1]{%
  \ifnum#1<10 \phantom{0}#1\else #1\fi
}

\usepackage[utf8]{inputenc} % allow utf-8 input
\usepackage[T1]{fontenc}    % use 8-bit T1 fonts
\usepackage{hyperref}       % hyperlinks
\usepackage{url}            % simple URL typesetting
\usepackage{booktabs}       % professional-quality tables
\usepackage{amsfonts}       % blackboard math symbols
\usepackage{nicefrac}       % compact symbols for 1/2, etc.
\usepackage{microtype}      % microtypography
\usepackage{xcolor}         % colors

\title{Aligning Few-Step Generative Models by\\
Amortizing Sample-based Variational Inference}

\makeatletter
\renewcommand\@fnsymbol[1]{\ensuremath{%
  \ifcase#1\or 1\or 2\or 3\or 4\or 5\or 6\or 7%
  \else\@ctrerr\fi}}
\makeatother

\author{%
  Jaewoo Lee\textsuperscript{*}\,\footnotemark[1]\;\;\footnotemark[2]
  \And Hyeongyu Kang\textsuperscript{*}\,\footnotemark[1]
  \And Dohyun Kim\footnotemark[1]
  \And Kyuil Sim\footnotemark[1]
  \AND Woocheol Shin\footnotemark[1]
  \And Minsu Kim\footnotemark[1]\;\;\footnotemark[3]
  \And Taeyoung Yun\footnotemark[1]
  \And Jeongjae Lee\footnotemark[1]
  \AND Sanghyeok Choi\footnotemark[4]
  \And Tabitha Edith Lee\footnotemark[3]\;\;\footnotemark[5]
  \And Jong Chul Ye\textsuperscript{\textdagger}\,\footnotemark[1]
  \And Jinkyoo Park\textsuperscript{\textdagger}\,\footnotemark[1]\;\;\footnotemark[6]
  \AND
  \textnormal{%
  \footnotemark[1]\,\;KAIST
  \quad
  \footnotemark[2]\,\;MongooseAI
  \quad
  \footnotemark[3]\,\;Mila -- Quebec AI Institute
  }\\
  \textnormal{%
  \footnotemark[4]\,\;University of Edinburgh
  \quad
  \footnotemark[5]\,\;Université de Montréal\quad
  \footnotemark[6]\,\;Omelet
  } \\\\
  \texttt{\{\href{mailto:jaewoo@kaist.ac.kr}{jaewoo}, \href{mailto:khg2000v@kaist.ac.kr}{khg2000v}\}@kaist.ac.kr}
}

\begin{document}
\maketitle

\begingroup
\renewcommand\thefootnote{}%
\footnotetext{%
  \textsuperscript{*}Equal contribution authors.\quad
  \textsuperscript{\textdagger}Corresponding authors.\\%
  \hspace*{2em}Project page: \url{https://jaewoopudding.github.io/fav_project_page}}
\endgroup
\vspace{-10pt}

\begin{abstract}
Aligning a few-step generative model is challenging, since existing alignment frameworks typically rely on restrictive assumptions: a tractable likelihood, a specific ODE/SDE solver, or a particular model family. We introduce \textbf{FAV} \textit{(\textbf{F}ew-step Generative Models \textbf{A}lignment via Sample-based \textbf{V}ariational Inference)}, a general alignment framework that requires only sample access to the generator and the reference distribution. We cast alignment as sampling from a reward-tilted distribution anchored to a reference distribution. We leverage Stein Variational Gradient Descent as a sample-based variational inference scheme and amortize its particle updates into the generator parameters via fixed-point regression. We evaluate FAV on two domains: robotics manipulation and image generator alignment. On generative policy alignment for robotic manipulation, FAV outperforms prevailing policy extraction baselines across 56 offline and 30 offline-to-online RL tasks. For image generator alignment, FAV fine-tunes diverse few-step backbones, including GAN, drifting model, consistency models, and flow maps, scaling from ImageNet-$256$ to 1024$^2$ text-to-image synthesis. Code is available at \href{https://github.com/Jaewoopudding/FAV}{\texttt{this link}}.
\end{abstract}

% \begin{figure}[h]
% \vspace{-0pt}
% \begin{center}
% \centerline{\includegraphics[width=0.5\textwidth]{Fig/main_rl.pdf}}
% \end{center}
\begin{figure}[h]
\vspace{-10pt}
    \centering
    \begin{minipage}{0.51\textwidth}
        \centering
        \includegraphics[width=\textwidth]{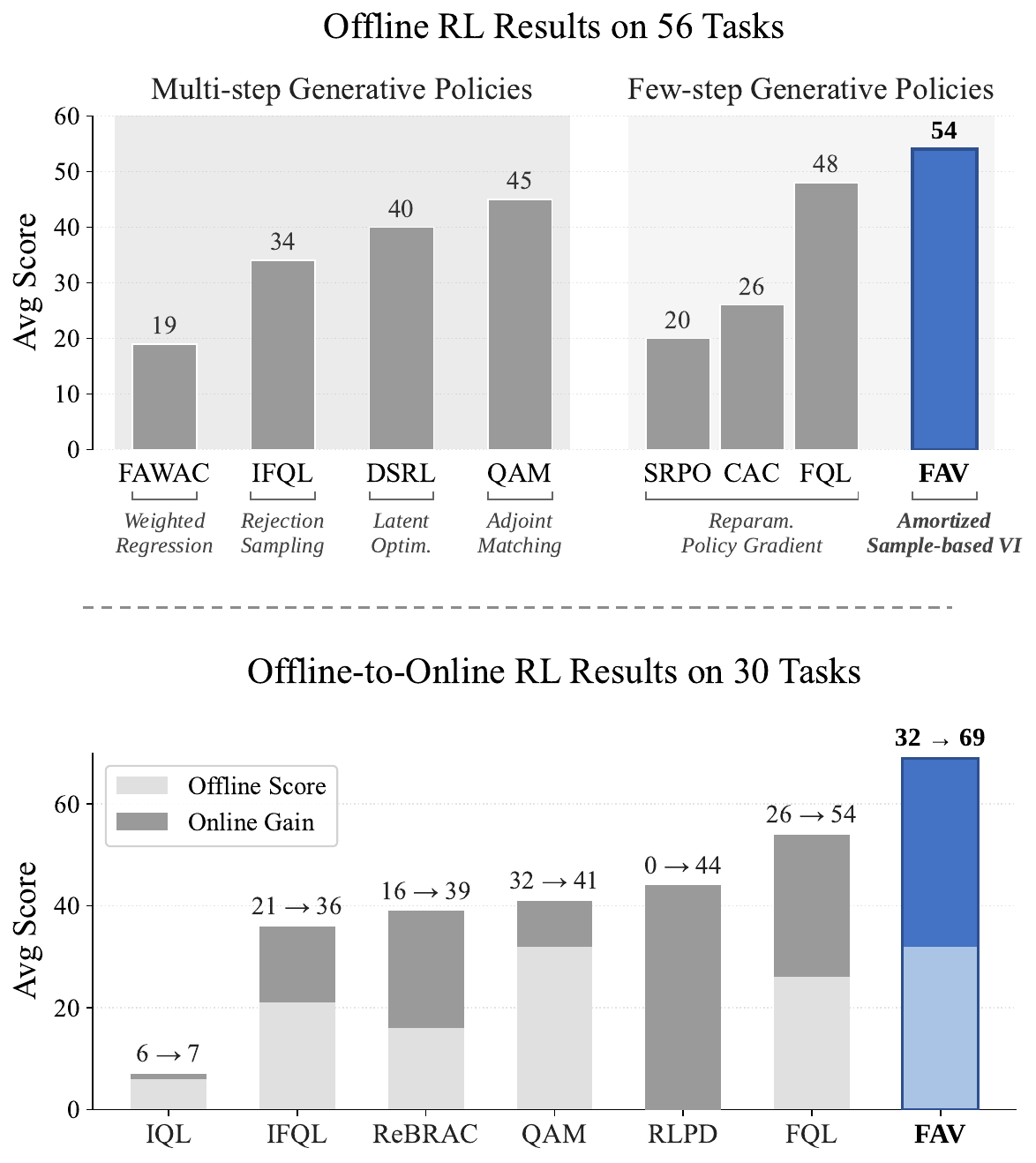}
    \end{minipage}
    \hfill
    \begin{minipage}{0.48\textwidth}
        \centering
        \includegraphics[width=\textwidth]{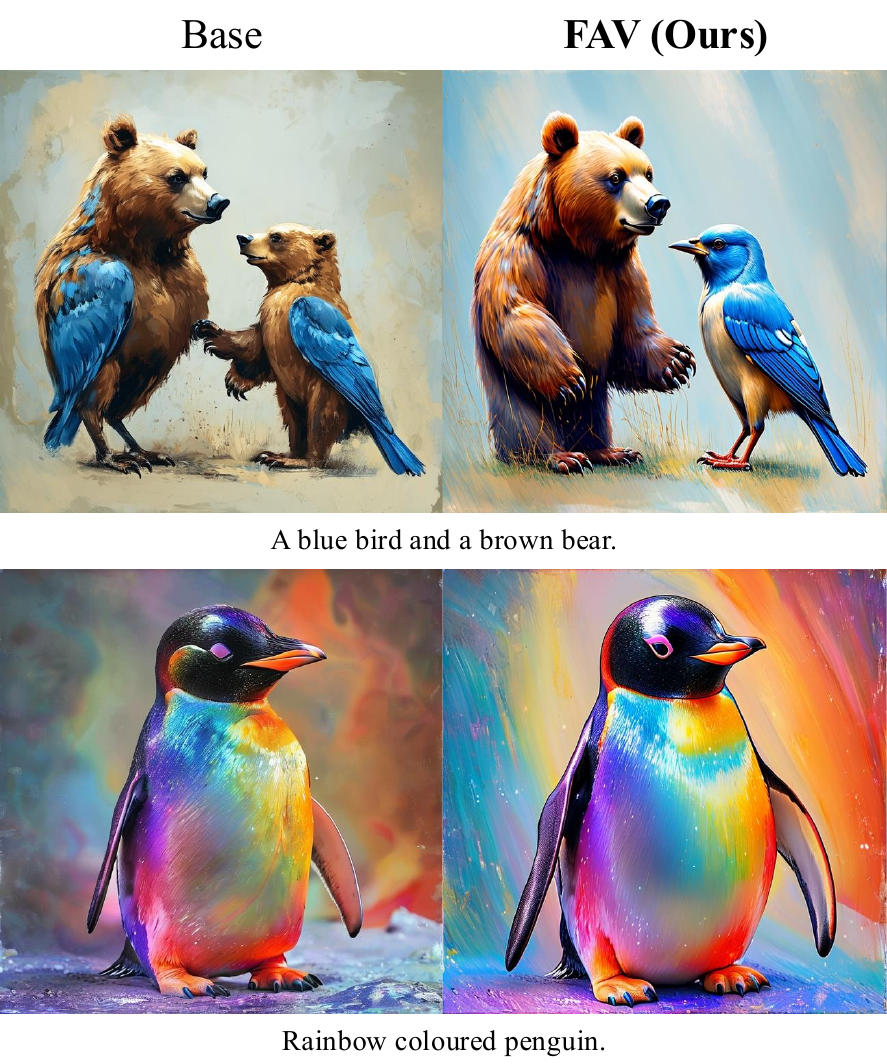}
    \end{minipage}
    \vspace{-0pt}
    \caption{\textit{Left}: Sampling from the Q-tilted distribution via FAV yields state-of-the-art performance on offline and offline-to-online RL tasks. \textit{Right}: Sampling from a human-preference-tilted distribution through FAV improves image quality of a high-resolution text-to-image generator.}
    \label{fig:main_fig}
    %\vspace{-30pt}
\end{figure}

\section{Introduction}
Recently emerging few-step generators~\citep{song2023consistency, fransone, gengmean, zhouinductive, boffiflow, deng2026generative} are implicit models: they do not admit tractable likelihood evaluation~\citep{ai2025joint}, and their generation procedures are often tied to specialized samplers~\citep{song2023consistency, kimconsistency, deng2026generative}. However, existing alignment methods for generative models rely on structural assumptions that few-step generators often do not satisfy: tractable likelihoods over denoising trajectories \citep{blacktraining, fan2023dpok, venkatraman2024amortizing, liu2023flow,lee2025diffusion,kang2025diffusion, wallace2024diffusion, li2024aligning, zhudspo}, rely on specific ODE/SDE solvers \citep{domingoadjoint, liuflow,xue2025dancegrpo}, or model-specific formulations \citep{kang2023efficient, ding2024diffusion, zhangenergy,xue2025advantage, zheng2025diffusionnft}. 

To address this limitation, we propose \textit{\textbf{F}ew-step Generative Models \textbf{A}lignment via Sample-based \textbf{V}ariational Inference}, \textbf{(FAV)}. Our key design principle is to make the alignment procedure sample-based, thereby decoupling it from the generator family and sampling dynamics. FAV formulates alignment as sampling from the reward-tilted distribution:
\begin{align*}
q^*(x)\propto p_\text{ref}(x)\exp(r(x)),
\end{align*}
where $p_{\mathrm{ref}}$ denotes a reference distribution, such as a distribution induced by a pretrained generator or an empirical data distribution. FAV approximates $q^*$ using Stein Variational Gradient Descent (SVGD)~\citep{liu2016stein} and amortizes the resulting SVGD transport into the generator through fixed-point regression~\citep{wang2016learning, deng2026generative}. Because SVGD requires the score of $p_{\mathrm{ref}}$, which is intractable for empirical data and implicit generators, we estimate it nonparametrically with kernel density estimation (KDE)~\citep{parzen1962estimation}. 
This construction requires only samples from the generative model and $p_{\mathrm{ref}}$, therefore does not require assumptions such as tractable likelihoods, explicit sampling trajectories, or a particular model family. As a result, FAV can fine-tune diverse few-step generators, including VAEs~\citep{kingma2014auto}, GANs~\citep{goodfellow2014generative}, consistency models~\citep{song2019generative, luo2023latent, lusimplifying}, and flow-map models~\citep{geng2025improved, boffiflow, boffibuild}. FAV can also be used for pre-training by choosing the empirical data distribution as $p_{\mathrm{ref}}$, yielding a sampler for the reward-tilted distribution without a pre-trained generator.

% This formulation has two practical benefits. First, the low per-sample cost of few-step generators makes sample-based alignment efficient. Second, the flexible choice of $p_{\mathrm{ref}}$ allows FAV to be used for both pre-training from data and fine-tuning from a pretrained generator.

%. To this end, we can flexibly set the reference distribution for FAV, supporting both pretraining and fine-tuning by setting either the empirical dataset distribution or a pre-trained model distribution.

% policy extraction for RL \citep{peters2007reinforcement, peng2019advantage, wu2019behavior, nair2020awac,tarasov2023revisiting, kang2023efficient, chenoffline, li2024aligning, ding2024diffusion, zhangenergy} and image generator alignment \citep{uehara2024fine, venkatraman2024amortizing, uehara2024understanding, uehara2025inference, domingoadjoint, kimtest, kang2025diffusion, lee2025diffusion}. 

We evaluate FAV on robotics manipulation and image generator alignment tasks. In robotics manipulation, FAV achieves state-of-the-art performance on 56 offline RL tasks and 30 offline-to-online RL tasks from OGBench \citep{parkogbench} and D4RL \citep{fu2020d4rl}, outperforming prevailing policy extraction paradigms such as reparameterized policy gradient \citep{park2024value, fql_park2025}. In image generator alignment, we fine-tune few-step generators at two scales: at ImageNet-256~\citep{deng2009imagenet}, we align StyleGAN-XL~\citep{sauer2022stylegan}, drifting model~\citep{deng2026generative}, inductive moment matching~\citep{zhouinductive}, and iMeanFlow~\citep{geng2025improved} to optimize an aesthetic reward~\citep{schuhmann2023laion}. We also generalize FAV to optimize a black-box reward by using zeroth-order gradient estimation. At the $1024^2$ resolution text-to-image scale, we align SANA-Sprint~\citep{chen2025sana} for a human preference score~\citep{wu2023human} and an NSFW safety detector \citep{falconsai2024nsfw}. Across both scales, FAV is agnostic to model architecture and sampler, and attains high reward while mitigating reward overoptimization.

\section{Related works}

\paragraph{Few-step generative models.} GAN~\citep{goodfellow2014generative, sauer2022stylegan} and VAE \citep{kingma2014auto} are the earliest class of single-step deep generative models, but are notoriously unstable to train or tend to generate blurry samples. Following the advent of diffusion and flow models~\citep{song2019generative, ho2020denoising, lipmanflow, liu2023flow}, consistency models~\citep{song2023consistency, songimproved, gengconsistency, luo2023latent} enable few-step generation by learning a consistency function that maps any point along a probability-flow ODE trajectory to its data point. Distillation-based approaches improve few-step diffusion generation by employing an adversarial objective \citep{sauer2024adversarial} or distribution matching objective \citep{yin2024one, luo2023diff}. Flow maps~\citep{kimconsistency, fransone, gengmean, boffiflow, boffibuild} instead learn the jump operator corresponding to ODE integration between two denoising steps, refining sample quality as the number of steps grows. More recently, drifting model~\citep{deng2026generative} learns one-step noise-to-data mappings by amortizing the transport induced by a kernel mean-shift drift field~\citep{cheng1995mean}. Despite this progress, pre-trained few-step generators can exhibit unintended behaviors, yet alignment methods tailored to few-step models remain underexplored.

\paragraph{Alignment of generative models.}
Generative model alignment can be pursued through two broad paradigms: maximizing the reward by seeking $\arg\max_{q_\theta} \mathbb E_{x\sim q_\theta}[r(x)]$, or sampling from the reward-tilted distribution $ p_{\mathrm{ref}}(x)\exp(r(x))$. For reward maximization, RL-based methods~\citep{blacktraining} and direct backpropagation approaches~\citep{clarkdirectly, prabhudesai2023aligning} have been proposed. However, these methods often suffer from severe reward overoptimization, leading to degraded sample quality and reduced diversity. Sampling from a reward-tilted distribution mitigates overoptimization by anchoring the optimization to the $p_\text{ref}$. To sample from the reward-tilted distribution, prior work has explored KL-regularized RL defined on diffusion denoising Markov decision process (MDP)~\citep{fan2023dpok, uehara2024fine, uehara2024understanding, liderivative, liuflow, kang2025diffusion, lee2025diffusion}, sampling-based methods~\citep{skreta2025feynman, kimtest, zhang2025inference}, guidance methods~\citep{chungdiffusion,lu2023contrastive,holderrieth2025glass, holderrieth2026diamond}, and stochastic optimal control~\citep{domingoadjoint, li2026q}. Despite the substantial success of reward-tilted sampling, extending this success to few-step generative models remains underexplored.% Despite this progress, extending alignment paradigms developed for diffusion and flow models to few-step generators remains fundamentally challenging. Existing methods typically rely on at least one of the following requirements: (1) explicit likelihood evaluation along a sequential denoising process~\citep{fan2023dpok, venkatraman2024amortizing, kimtest}; (2) iterative guidance throughout the generation trajectory~\citep{chungdiffusion, holderrieth2025glass, holderrieth2026diamond}; (3) a specific SDE/ODE solver or sampling dynamics~\citep{domingoadjoint, liuflow}; or (4) a particular model family~\citep{zhangenergy, xue2025advantage, zheng2025diffusionnft}. These restrictions call for an alignment method that can align few-step generative models without relying on restrictive assumptions.

\section{Preliminaries}

\subsection{Alignment as sampling from a reward-tilted distribution}
Given a reference distribution $p_{\text{ref}}(x)$ and an objective function $r(x)$ (e.g., a reward model or Q-function), we cast the alignment problem as sampling from the unnormalized density:
\begin{align}
\label{eq:tilted distribution}
q^*(x)  \propto p_{\text{ref}}(x) \exp\left(\beta \cdot r(x)\right),
\end{align}
where $\beta>0$ is a temperature parameter governing the strength of the alignment \citep{kimtest}. This formulation naturally emerges from Variational Inference (VI), which is equivalent to the KL-regularized reinforcement learning objective \citep{jaques2019way,wu2019behavior,korbak2022rl, fan2023dpok}. The optimal variational distribution $q^*$ is the minimizer of the following KL objective:
\begin{align}
\label{eq: optimal distribution of VI}
q^*(x) = \arg\min_{q} D_{\text{KL}}\left(q(x) \,\middle\|\,  \frac{1}{Z}p_{\text{ref}}(x) \exp\left(\beta\cdot{r(x)}\right)\right),
\end{align}
where the $Z$ is a normalizing constant. Sampling from $q^*(x)$ can yield high-reward samples while preserving the property of the reference distribution \citep{uehara2024understanding, uehara2025inference}. This perspective has led to two primary application domains: conservative policy extraction in Reinforcement Learning (RL) and reward alignment of image generators. In offline RL, approximating $q^*(x)$ via weighted behavior cloning \citep{kang2023efficient, ding2024diffusion, zhangenergy}, GFlowNet \citep{venkatraman2024amortizing}, guidance \citep{lu2023contrastive}, and Adjoint Matching \citep{li2026q} have shown strong performance on challenging robotics manipulation tasks. In image generation, reward-weighted flow matching \citep{xue2025advantage, zheng2025diffusionnft}, entropy-regularized RL \citep{wallace2024diffusion,liuflow,uehara2024fine,kang2025diffusion}, stochastic optimal control \citep{domingoadjoint}, and probabilistic inference \citep{kimtest, lee2025diffusion} have been successfully applied to align image generators.

\subsection{Stein Variational Gradient Descent}
Standard VI objectives typically require density evaluation of $q(x)$ \citep{rezende2015variational, blei2017variational}, which is intractable when $q(x)$ is induced by an implicit generator. SVGD~\citep{liu2016stein,liu2017stein} bypasses this requirement by representing $q(x)$ with particles and iteratively transporting them toward the target distribution $p(x)$:
\begin{align}
\label{eq: SVGD update}
x_i^{\ell+1} \leftarrow x_i^{\ell} + \epsilon \phi_{q_\ell,p}^*(x_i), \quad \forall i = 1, \cdots, n,
\qquad
\phi_{q_\ell, p}^* = \arg \max_{\phi\in \mathcal F}
\left\{
-\left.\frac{d}{d\epsilon}\mathrm{KL}(q_{\ell} \,\|\, p)\right|_{\epsilon=0}
\right\}.
\end{align}
Here, $q_{\ell}$ denotes the empirical distribution of the particles at the $\ell$-th iteration, $\epsilon$ denotes the step size, and $\phi_{q_\ell, p}^*(x)$ represents the optimal Stein velocity field that maximizes the decreasing rate of the KL divergence between $q_\ell$ and the target $p$. When $\mathcal {F}$ is restricted to be the unit ball of a reproducing kernel Hilbert space (RKHS) associated with a positive definite kernel $k(x,x')$, the gradient of the KL divergence in~\Cref{eq: SVGD update} leads to the closed-form solution \citep{liu2016stein}:
\begin{align}
\label{optimal transport vector}
\phi_{q_{\ell},p}^*(\cdot)\propto \mathbb{E}_{x\sim q_{\ell}}\!\left[\mathcal{A}_pk(x,\cdot)\right]
= \mathbb{E}_{x\sim q_{\ell}}\!\left[k(x,\cdot)\,\nabla_x \log p(x)+\nabla_x k(x,\cdot)\right],
\end{align}
where $\mathcal {A}_p$ is a linear operator called Stein operator. Therefore, SVGD iteratively pushes particles using the optimal gradient direction $\phi^*$ and this eventually converge to the target $p$ with sufficiently small $\epsilon$, under which $\phi^*_{q_{\infty},p} \equiv 0$ (i.e. $q_{\infty} = p$ if and only if $\phi^*_{q_{\infty},p} \equiv 0$ when $k(x,x')$ is strictly positive definite in an appropriate sense \citep{liu2016kernelized, oates2017control, chwialkowski2016kernel} such as the RBF kernel). % By setting $q(x)$ as the distribution of the implicit generator, SVGD requires only samples from $q(x)$, rather than tractable likelihood evaluation of $q(x)$.

\section{Challenges in aligning few-step generative models}

Aligning few-step generative models has a key desideratum: improving sample quality while preserving fast generation. Under this advantage, few-step generative policy alignment can yield expressive policies with single-step inference, avoiding backpropagation through time and the credit-assignment challenges of multi-step generative policies. In image generator alignment, it enables high-reward generation without sacrificing the efficiency of few-step generators. However, existing alignment methods do not naturally extend to few-step generative models. To clarify this limitation, we categorize few-step generators into two classes: \textbf{(i) one-step noise-to-data mappings} and \textbf{(ii) flow maps}, and examine how the existing methods break down in each case.

\paragraph{One-step noise-to-data mappings.}
VAE \citep{kingma2014auto}, GAN \citep{goodfellow2014generative}, and Drifting Models \citep{deng2026generative} map latent noise directly to data in a single feed-forward pass. They do not expose the denoising trajectory required by prior alignment methods: RL-based approaches assume a sequential denoising process \citep{fan2023dpok, blacktraining, liuflow} or a flow matching backbone \citep{xue2025advantage, zheng2025diffusionnft}, whereas SOC-based methods assume explicit ODE/SDE dynamics \citep{domingoadjoint, liuvalue}. Moreover, directly measuring the KL divergence for regularization may require additional effort because it requires explicit likelihood evaluation.

\paragraph{Flow maps.} Recent emergent flow map generators \citep{fransone, zhouinductive, boffiflow, boffibuild, gengmean}, including consistency models \citep{song2023consistency, kimconsistency, lusimplifying} learn to jump between two noise levels. Flow map inference is performed using learned jump operators between two timesteps rather than numerically integrating the underlying ODE or SDE. Specifically, multistep consistency sampling, adopted by consistency models \citep{song2023consistency, luo2023latent, songimproved, lusimplifying, gengconsistency} and distilled diffusion models \citep{sauer2024adversarial, yin2024one, yin2024improved}, generates samples through a discrete iteration of consistency-map evaluations and perturbation with i.i.d. sampled Gaussian noise. Therefore, applying alignment methods built upon SDE/ODE solvers is not straightforward for flow map models.
% These limitations motivate our sample-based formulation, which only requires samples from the generator and the reference distribution. We next instantiate this principle by using SVGD to transport generator samples toward the reward-tilted distribution and amortizing the resulting transport into the generator.

\section{Aligning few-step generative models via sample-based VI}
In this section, we introduce FAV, a sample-based alignment framework for few-step generative models. Our key design principle is to build a sample-based method, making the alignment process agnostic to the generator family and sampling procedure. A sample-based approach is well-suited to few-step generators, whose low per-sample cost makes sample-based alignment efficient. Our method consists of three steps: (i) approximating the reward-tiled distribution via sample-based SVGD, (ii) estimating the intractable reference score through KDE, and (iii) amortizing the resulting Stein transport into the generator to preserve fast generation.

\subsection{Stein velocity field towards tilted distribution.} Let $f_\theta:\mathcal Z\rightarrow \mathcal X$ be a neural network parameterized mapping function that transforms a latent variable $z$ drawn from a prior $\mathcal Z$ into the data sample $x$ of $\mathcal{X}$, and $q_\theta$ be the empirical distribution of $f_\theta(\epsilon)$ with $\epsilon\sim p_\text{noise}$. SVGD offers a sample-based method for approximating the reward-tilted distribution $q^*(x)$ by only requiring samples from the implicit generator $f_\theta(\epsilon)$. By substituting the target distribution with the reward-tilted distribution, we derive the optimal Stein velocity field $\phi^*_{q_\theta,q^*}$ that drives samples from the current model $q_\theta$ toward $q^*$:
\begin{align}
\label{eq: optimal transport vector for reward tilted distribution}
\phi^*_{q_\theta, q^*}(x) 
&= \mathbb{E}_{x' \sim q_\theta} \Big[ \underbrace{k(x', x) \nabla_{x'} \log p_{\text{ref}}(x')}_{\text{Prior Alignment}} + \underbrace{\beta \cdot k(x', x) {\nabla_{x'} r(x')}{}}_{\text{Reward Guidance}} + \underbrace{\nabla_{x'} k(x', x)}_{\text{Diversity Enforcement}} \Big]. 
\end{align}
Note that the reward gradient can be computed directly from a differentiable reward function, or estimated using zeroth-order methods for black-box rewards. See \Cref{App:FAV with non-differentiable reward} for details about zeroth-order gradient estimation.

\subsection{KDE for intractable score estimation} 
\label{KDE for Intractable Score Estimation}
Since $\nabla \log p_{\mathrm{ref}}$ is typically intractable, we estimate it nonparametrically via kernel density estimation (KDE)~\citep{li2018gradient, zhou2020nonparametric, song2019generative}. Instantiating both the KDE and SVGD kernel to be Gaussian RBFs $k_\sigma$ with shared bandwidth $\sigma$ yields the following approximated transport field:
\begin{align}
\label{eq: approximated stein velocity field}
\hat{\phi}_{q_\theta, q^*_\sigma}^*(x) = \mathbb{E}_{\substack{x' \sim q_\theta \\ x^{\text{ref}} \sim p_{\text{ref}}}} \left[k_\sigma(x', x) \left( \frac{\tilde{k}_\sigma(x', x^{\text{ref}})(x^{\text{ref}} - x')}{ \sigma^2} + \beta\cdot{\nabla_{x'} r(x')}{} + \frac{(x - x')}{ \sigma^2} \right) \right],
\end{align}
where $\tilde{k}_\sigma(x', x^{\text{ref}}) := \frac{k_\sigma(x', x^{\text{ref}})}{\mathbb E_{\bar x^\text{ref}}[k_\sigma(x',\bar x^\text{ref})]}$. The resulting velocity field $\hat{\phi}^*_{q_\theta, q^*_\sigma}$ pushes samples toward a surrogate posterior $q^*_\sigma(x) \propto p_\sigma(x)\exp(\beta\cdot r(x))$, where $p_\sigma = p_{\text{ref}} * k_\sigma$. We establish the consistency conditions under which $q^*_\sigma \to q^*$ and a detailed derivation of \Cref{eq: approximated stein velocity field} in \Cref{App: Score estimation through Gaussian kernel}. The choice of $p_{\mathrm{ref}}$ determines the training regime. When $p_{\mathrm{ref}}$ is the empirical data distribution, FAV directly learns an aligned generator from data without a separate reference generator. When $p_{\mathrm{ref}}$ is the distribution of a pre-trained model, FAV fine-tunes that generator.

% Note that we optimize towards a smoothed . KDE is asymptotically consistent as the $N\sigma \to \infty$ and $\sigma \to 0$, where $N$ is the number of samples.

\definecolor{codepink}{RGB}{220,40,130}
\definecolor{codeteal}{RGB}{55,125,130}

\captionsetup[algorithm]{
  labelfont=bf,
  textfont=normalfont,
  labelsep=period
}

\lstdefinestyle{favalg}{
  language=Python,
  basicstyle=\ttfamily,
  columns=fullflexible,
  keepspaces=true,
  showstringspaces=false,
  frame=none,
  aboveskip=0.6em,
  belowskip=-0.8em,
  commentstyle=\color{codeteal},
  keywordstyle=\color{codepink},
  morekeywords={
    generator,
    randn,
    sample_from,
    get_stein_transport,
    stopgrad,
    mse_loss
  },
  escapeinside={(*@}{@*)}
}

\begin{figure}[t]
    \centering
    \begin{minipage}[t]{0.55\textwidth}
        \vspace{-10pt}
        \begin{algorithm}[H]
        \caption{\textbf{FAV Loss.}}
        \label{alg:fav_training_loss}
        \begin{lstlisting}[style=favalg]
# p_ref  : dataset / pre-trained model
# r      : reward function

eps      = randn([N_gen, C])
x_ref    = sample_from(p_ref, N_ref) 
x        = generator(eps)    

phi      = get_stein_transport(x, x_ref, r)
x_target = stopgrad(x + phi)

loss     = mse_loss(x, x_target)
        \end{lstlisting}
        \end{algorithm}
    \end{minipage}
    \hfill
    \begin{minipage}[t]{0.40\textwidth}
        \vspace{0pt}
        \centering
        \includegraphics[width=\linewidth]{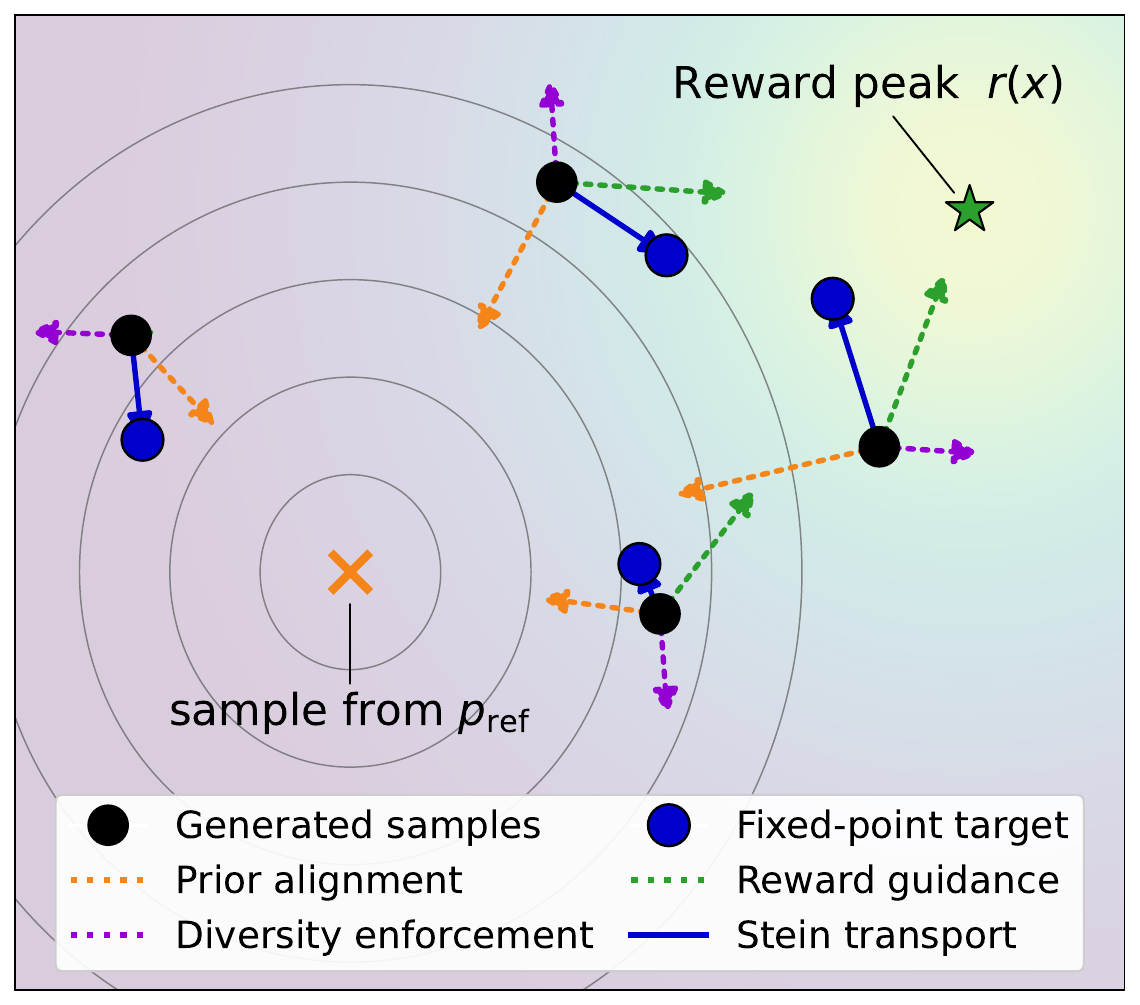}
        \captionof{figure}{Illustration of SVGD transport.}
        \label{fig:illustrative sample}
    \end{minipage}
\end{figure}

\subsection{Amortization via fixed-point regression.} While iterative SVGD updates of the form $ x'\leftarrow x +\hat\phi_{q_\theta,q^*_\sigma}^*(x)$ in  \Cref{eq: approximated stein velocity field} can approximate the reward-tilted distribution, repeated transport incurs inference-time overhead that erodes the core advantage of few-step generative models: fast sampling. Therefore, we amortize the transport via the Stein velocity field into the network parameters $\theta$ through a fixed-point regression \citep{wang2016learning, deng2026generative, lai2026unified}:
\begin{align}
\label{eq: FAV loss function}
\mathcal L(\theta)&=\mathbb E_{\epsilon\sim \mathcal N}[||\text{stopgrad}(f_\theta(\epsilon)+\hat\phi^*_{q_\theta,q^*_\sigma }(f_\theta(\epsilon)))-f_\theta(\epsilon)||^2_2].
\end{align} 
The stop-gradient operator casts the transported sample as a stable regression target. Minimizing this objective distills the SVGD update into the generator, moving the model distribution $q_\theta$ toward the smoothed reward-tilted target $q_\sigma^*$ while preserving the advantage of few-step sampling.

% \paragraph{Deep Kernel \citep{wilson2016deep} for High-dimensional Data.} 
% For high-dimensional data, such as images, raw spatial distances often fail to capture similarity between data points \citep{hadsell2006dimensionality, zhang2018unreasonable}. We therefore define the kernel in a learned representation space induced by a fixed pre-trained encoder 
% $\psi$, trained in a self-supervised manner \citep{radford2021learning}. Then the distance between data points denotes the meaningful proximity. Our representation-space amortization objective is given by:
% \begin{align} \label{eq:deep kernel FAV loss}
% \mathcal L(\theta)&=\mathbb E_{\epsilon\sim \mathcal N}[||\text{stopgrad}(\psi(f_\theta(\epsilon))+\hat\phi^*_{q_\theta,p }(\psi(f_\theta(\epsilon))))-\psi(f_\theta(\epsilon))||^2_2].
% \end{align}

\subsection{FAV in pre-trained representation space} 
For high-dimensional data such as images, kernels on raw spatial distances often fail to capture perceptually meaningful similarity between data points~\citep{hadsell2006dimensionality, zhang2018unreasonable}. 
Motivated by kernel-based distribution matching in learned representation spaces~\citep{binkowski2018demystifying, deng2026generative}, we perform FAV training in the representation space of the pre-trained encoder $\psi$ rather than in the raw space. Our amortization objective on the representation space is given by:
\begin{align} \label{eq:deep kernel FAV loss}
\mathcal L(\theta)&=\mathbb E_{\epsilon\sim \mathcal N}[||\text{stopgrad}(\psi(f_\theta(\epsilon))+\hat\phi^*_{q_\theta,q^*_\sigma }(\psi(f_\theta(\epsilon))))-\psi(f_\theta(\epsilon))||^2_2].
\end{align}

% \paragraph{Theoretical Analysis.}
% FAV replaces the intractable reference density $p_{\text{ref}}$ with its KDE estimate, thereby yielding a surrogate-optimal policy $q_\sigma^*(x)$. Our analysis quantifies the gap between $q_\sigma^*(x)$ and the true optimal policy $q^*(x)$. In particular, we show that the degradation of $q_\sigma^*(x)$ comes from two effects of KDE: the smoothing bias induced by kernel convolution and the finite-sample estimation error induced by approximating the smoothed density with a finite number of samples. We further show that as the bandwidth $\sigma \to 0$ and the sample size $N \to \infty$, the degradation vanish and the bound asymptotically recovers the ideal improvement of $q^*(x)$. (See detail in Appendix \ref{Theoretical Analysis})

\section{Experiments}
In this section, we evaluate FAV on three domains: 2D toy setup, robotics manipulation, and image generator alignment. Through these experiments, we demonstrate that FAV is a domain-agnostic and powerful method for sampling from reward-tilted distributions, applicable from low-dimensional data, proprioceptive data, to high-resolution 1024$^2$ images. Moreover, we demonstrate that FAV generalizes across diverse classes of few-step generators, including single-step mappings such as VAE and GAN, as well as consistency models and flow maps. We provide the ablation and sensitivity analysis in \Cref{App: Ablation Analysis} and \Cref{App: Sensitivity Analysis}, respectively.

\subsection{Toy setting}
\label{toy setting}
\begin{figure}[t]
    \centering
    \includegraphics[width=\textwidth]{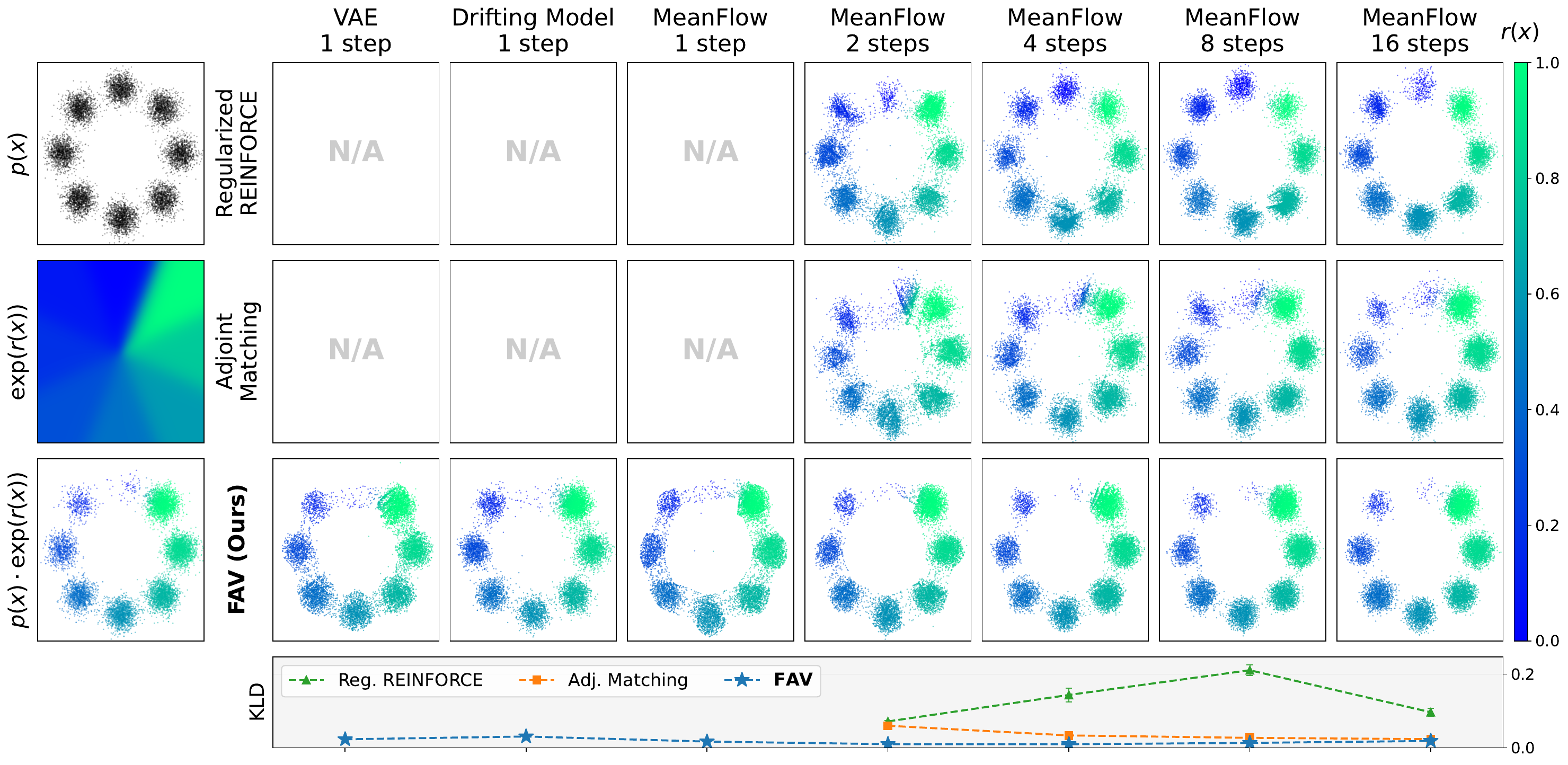}
    \caption{\textbf{Sampling from a reward-tilted distribution.} On the 8 Gaussian $p(x)$ with the reward function shown on the left, we target to sample from $p(x)\exp(r(x))$. Regularized REINFORCE and Adjoint Matching are not applicable to 1-step generators, whereas FAV applies uniformly across all architectures. For MeanFlow from 2 to 16 sampling steps, FAV consistently yields samples that better match the reward-tilted target and achieves lower KL divergence than baselines.}
    \label{fig:toy_result}
    \vspace{-1em}
\end{figure}

To build intuition for the ability of FAV to sample from reward-tilted distributions, we conduct a toy study on an 8-mode 2D Gaussian mixture. First, we pre-train three representative few-step generators: VAE \citep{kingma2014auto}, drifting model \citep{deng2026generative}, and MeanFlow \citep{gengmean}. Then, we fine-tune each model with FAV and two prevailing alignment baselines: policy gradient \citep{sutton1999policy} on a denoising MDP \citep{blacktraining} and Adjoint Matching \citep{domingoadjoint}.
Following~\citep{venkatraman2024amortizing, deleu2025relative}, we regularize the policy gradient with an approximate KL penalty to sample from the tilted distribution. Because baselines require SDE integration for sampling, we treat the average velocity of MeanFlow as an instantaneous velocity and inject noise at each denoising step. See \Cref{app: toy experiments} for details.

As shown in \Cref{fig:toy_result}, FAV consistently attains the lowest KL divergence to the target distribution across all settings. Policy gradient and Adjoint Matching are either inapplicable or non-trivial to adapt to the VAE, Drifting Model, and 1-step MeanFlow. Notably, both baselines degrade significantly in the few-step regime since they rely on alignment signals defined over multi-step denoising trajectories, which become ill-suited when the sampling process is collapsed into only a few steps. Overall, these results demonstrate that FAV is a strong, model-agnostic choice for aligning few-step generators.

\subsection{Aligning generative policy with FAV}
Sampling actions from the Q-tilted distribution: $p_{\mathrm{data}}(a \mid s)\exp(Q(s,a))$ is a promising approach for offline RL and offline-to-online RL \citep{peng2019advantage, wu2019behavior, nair2020awac, venkatraman2024amortizing, zhangenergy}. Our RL experiments assess two aspects of FAV: \textbf{(1) policy extraction capability} relative to prior policy extraction methods, and \textbf{(2) transition ability from offline to online learning}.

\subsubsection{Experimental setup}

\paragraph{Benchmarks.}
We evaluate FAV on 56 offline RL tasks: 50 reward-based variants (\texttt{-singletask}) from OGBench~\citep{parkogbench}, spanning 10 environments with diverse embodiments and control challenges, and 6 challenging antmaze tasks from D4RL~\citep{fu2020d4rl}. For offline-to-online RL, we select the 30 OGBench tasks across six environments on which FAV's offline performance is weakest. We run 8 seeds and report the mean and the 95\% confidence interval.

\paragraph{FAV configuration.} We instantiate FAV as a single-step noise-to-action generator $\pi_\theta(a \mid s, z): \mathbb{R}^s \times \mathbb{R}^a \rightarrow \mathbb{R}^a$, avoiding backpropagation through time and simplifying credit assignment relative to multi-step generative policies. We set an offline dataset as the reference distribution $p_\text{data}$ to pre-train the agent directly on offline data. Additionally, we propose \textbf{FAV-Adaptive}, which sets $\sigma$ automatically via Scott's rule \citep{scott2015multivariate}. We adopt the same actor and critic architectures as in FQL \citep{fql_park2025} and use the same optimization hyperparameters across all baselines to ensure a fair comparison.

\paragraph{Baselines.}
We compare FAV against 10 representative policy extraction methods in both offline RL and offline-to-online RL. For offline RL, we consider four categories of baselines: \textbf{(1) Gaussian policies.} IQL~\citep{kostrikovoffline} and ReBRAC~\citep{tarasov2023revisiting}. In particular, ReBRAC targets the Q-tilted distribution and produces actions in a single forward pass, analogous to FAV. \textbf{(2) Latent-space policies.} DSRL~\citep{wagenmakersteering}, which acts in the latent space of a pre-trained flow policy. While it is agnostic to policy type, it introduces additional inference-time overhead. \textbf{(3) Multi-step flow policies.} We compare FAV with FAWAC, IFQL \citep{hansen2023idql}, and QAM \citep{li2026q}, all of which require iterative denoising. Each of them extracts policy from value function via weighted behavioral cloning \citep{nair2020awac}, rejection sampling, and Adjoint Matching \citep{domingoadjoint}, respectively. \textbf{(4) Few-step distilled policies.} SRPO \citep{chenscore}, CAC \citep{ding2024diffusion}, and FQL \citep{fql_park2025} distill a generative model into a one- or few-step model: SRPO~\citep{chenscore} distills a diffusion score, CAC applies consistency training~\citep{song2023consistency}, and FQL distills from flow-matching behavioral policy. For offline-to-online RL, we additionally include RLPD~\citep{ball2023efficient}, a dedicated offline-to-online method. %, alongside IQL, ReBRAC, FQL, IFQL, and QAM.

\subsubsection{Results}

\begin{table*}[t]
\centering
\small
\setlength{\tabcolsep}{4pt}
\caption{\textbf{Offline RL performance.} FAV achieves the best overall average performance among all compared methods across 56 tasks from OGBench \cite{parkogbench} and D4RL \cite{fu2020d4rl}. FAV-Adaptive, with pre-calculated bandwidth through data, also outperforms all baselines in average performance.}
\label{table:main_offline_rl}
\resizebox{\linewidth}{!}{%
\begin{tabular}{lccccccccccc}
\toprule[1pt]
& \multicolumn{2}{c}{Gaussian Policies}
& \multicolumn{1}{c}{Latent Opt.}
& \multicolumn{3}{c}{Flow-based Policies}
& \multicolumn{3}{c}{Distillation Policies}
& \multicolumn{2}{c}{\textbf{Ours}} \\
\cmidrule(lr){2-3} \cmidrule(lr){4-4} \cmidrule(lr){5-7} \cmidrule(lr){8-10} \cmidrule(lr){11-12}
Task Categories (offline 1M)
& IQL & ReBRAC & DSRL & FAWAC & IFQL & QAM & SRPO & CAC & FQL & \textbf{FAV-Adaptive} & \textbf{FAV} \\
\midrule[1pt]
OGBench antmaze-large-navigate-singletask (5 tasks) 
& 53 \std{3} & 81 \std{5} & 40 \std{29} & 6 \std{1} & 28 \std{5} & 77 \std{5} & 11 \std{4} & 33 \std{4} & 79 \std{3} & 80 \std{2} & \textbf{87 \std{2}} \\
OGBench antmaze-giant-navigate-singletask (5 tasks) 
& 4 \std{1} & \textbf{26 \std{8}} & 0 \std{0} & 0 \std{0} & 3 \std{2} & 3 \std{1} & 0 \std{0} & 0 \std{0} & 9 \std{6} & \textbf{26 \std{3}} & \textbf{26 \std{5}} \\
OGBench humanoidmaze-medium-navigate-singletask (5 tasks) 
& 33 \std{2} & 22 \std{8} & 34 \std{20} & 19 \std{1} & 60 \std{14} & 63 \std{2} & 1 \std{1} & 53 \std{8} & 58 \std{5} & 44 \std{8} & \textbf{64 \std{10}} \\
OGBench humanoidmaze-large-navigate-singletask (5 tasks) 
& 2 \std{1} & 2 \std{1} & \textbf{10 \std{12}} & 0 \std{0} & \textbf{11 \std{2}} & 4 \std{3} & 0 \std{0} & 0 \std{0} & 4 \std{2} & 4 \std{1} & 5 \std{3} \\
OGBench antsoccer-arena-navigate-singletask (5 tasks) 
& 8 \std{2} & 0 \std{0} & 28 \std{9} & 12 \std{0} & 33 \std{6} & \textbf{61 \std{1}} & 1 \std{0} & 2 \std{4} & 60 \std{2} & 60 \std{2} & 59 \std{2} \\
OGBench cube-single-play-singletask (5 tasks) 
& 83 \std{3} & 91 \std{2} & 93 \std{14} & 81 \std{4} & 79 \std{2} & 57 \std{12} & 80 \std{5} & 85 \std{9} & \textbf{96 \std{1}} & 92 \std{1} & 93 \std{1} \\
OGBench cube-double-play-singletask (5 tasks) 
& 7 \std{1} & 12 \std{1} & \textbf{53 \std{14}} & 5 \std{2} & 14 \std{3} & 30 \std{7} & 2 \std{1} & 6 \std{2} & 29 \std{2} & 24 \std{3} & 26 \std{2} \\
OGBench scene-play-singletask (5 tasks) 
& 28 \std{1} & 41 \std{3} & \textbf{88 \std{9}} & 30 \std{3} & 30 \std{3} & 59 \std{0} & 20 \std{1} & 40 \std{7} & 56 \std{2} & 55 \std{1} & 55 \std{1} \\
OGBench puzzle-3x3-play-singletask (5 tasks) 
& 9 \std{1} & 21 \std{1} & 0 \std{0} & 6 \std{2} & 19 \std{1} & 19 \std{7} & 18 \std{1} & 19 \std{0} & 30 \std{1} & 63 \std{8} & \textbf{73 \std{8}} \\
OGBench puzzle-4x4-play-singletask (5 tasks)
& 7 \std{1} & 14 \std{1} & 37 \std{13} & 1 \std{0} & 25 \std{5} & \textbf{38 \std{3}} & 10 \std{3} & 15 \std{3} & 17 \std{2} & 12 \std{2} & 16 \std{5} \\

D4RL antmaze (6 tasks)
& 57 & 78 & 56 \std{2} & 44 \std{3} & 65 \std{7} & 80 \std{10} & 74 & 30 \std{3} & \textbf{84 \std{3}} & 79 \std{1} & 80 \std{6} \\
\midrule[1pt]
Total Average (56 tasks)
& 27 & 36 & 40 & 19 & 34 & 45 & 20 & 26 & 48 & 50 & \textbf{54} \\
\bottomrule[1pt]
\end{tabular}%
}
\vspace{-5pt}
\end{table*}

\begin{table*}[t]
\centering
\small
\setlength{\tabcolsep}{4pt}
\caption{\textbf{Offline-to-online RL performance.} We consider the 30 OGBench tasks from the six environments where FAV attains the lowest average performance. After an additional 1M online steps, FAV achieves the best post-training performance.}
\label{tab: offline_to_online_rl}
\resizebox{\linewidth}{!}{%
\begin{tabular}{lccccccc}
\toprule[1pt]
Task Categories (offline 1M $\rightarrow$ online 1M)
& IQL & ReBRAC & RLPD & IFQL & QAM & FQL & FAV \\
\midrule[1pt]
antmaze-giant-navigate-singletask (5 tasks)
& 4 \std{2} $\rightarrow$ 3 \std{2}
& 32 \std{7} $\rightarrow$ \textbf{99} \std{1}
& 0 \std{0} $\rightarrow$ 64 \std{15}
& 2 \std{2} $\rightarrow$ 0 \std{0}
& 4 \std{1} $\rightarrow$ 11 \std{6}
& 0 \std{0} $\rightarrow$ 44 \std{10}
& 28 \std{5} $\rightarrow$ 74 \std{3} \\
humanoidmaze-large-navigate-singletask (5 tasks)
& 2 \std{1} $\rightarrow$ 2 \std{1}
& 1 \std{2} $\rightarrow$ 0 \std{0}
& 0 \std{0} $\rightarrow$ 0 \std{0}
& 12 \std{2} $\rightarrow$ 12 \std{4}
& 4 \std{3} $\rightarrow$ \textbf{21} \std{7}
& 5 \std{2} $\rightarrow$ 4 \std{6}
& 6 \std{3} $\rightarrow$ 8 \std{6} \\
antsoccer-arena-navigate-singletask (5 tasks)
& 8 \std{3} $\rightarrow$ 4 \std{1}
& 0 \std{0} $\rightarrow$ 0 \std{0}
& 0 \std{0} $\rightarrow$ 37 \std{4}
& 27 \std{11} $\rightarrow$ 49 \std{10}
& 61 \std{4} $\rightarrow$ 93 \std{5}
& 57 \std{5} $\rightarrow$ 89 \std{5}
& 62 \std{4} $\rightarrow$ \textbf{92} \std{1} \\
cube-double-play-singletask (5 tasks)
& 2 \std{1} $\rightarrow$ 0 \std{0}
& 6 \std{2} $\rightarrow$ 35 \std{9}
& 0 \std{0} $\rightarrow$ 2 \std{2}
& 9 \std{1} $\rightarrow$ 54 \std{6}
& 26 \std{11} $\rightarrow$ 24 \std{21}
& 27 \std{2} $\rightarrow$ 75 \std{5}
& 25 \std{5} $\rightarrow$ \textbf{91} \std{2} \\
scene-play-singletask (5 tasks)
& 21 \std{3} $\rightarrow$ 33 \std{6}
& 41 \std{4} $\rightarrow$ 60 \std{0}
& 0 \std{0} $\rightarrow$ 59 \std{0}
& 47 \std{5} $\rightarrow$ 57 \std{2}
& 59 \std{1} $\rightarrow$ 60 \std{0}
& 55 \std{2} $\rightarrow$ 60 \std{0}
& 54 \std{2} $\rightarrow$ \textbf{62} \std{5} \\
puzzle-4x4-play-singletask (5 tasks)
& 2 \std{1} $\rightarrow$ 0 \std{0}
& 13 \std{2} $\rightarrow$ 38 \std{7}
& 0 \std{0} $\rightarrow$ \textbf{100} \std{0}
& 29 \std{4} $\rightarrow$ 45 \std{9}
& 38 \std{4} $\rightarrow$ 40 \std{17}
& 12 \std{3} $\rightarrow$ 50 \std{10}
& 18 \std{5} $\rightarrow$ 85 \std{9} \\
\midrule[1pt]
Total Average (30 tasks)
& 6 \std{1} $\rightarrow$ 7 \std{1}
& 16 \std{1} $\rightarrow$ 39 \std{2}
& 0 \std{0} $\rightarrow$ 44 \std{3}
& 21 \std{2} $\rightarrow$ 36 \std{3}
& 32 \std{2} $\rightarrow$ 41 \std{6}
& 26 \std{1} $\rightarrow$ 54 \std{3}
& 32 \std{2} $\rightarrow$ \textbf{69} \std{3} \\
\bottomrule[1pt]
\end{tabular}%
}
\label{tab:off2on_final}
\vspace{-15pt}
\end{table*}

\paragraph{FAV outperforms baselines on both Offline and Offline-to-Online RL.}
\Cref{table:main_offline_rl} reports results on 56 offline RL tasks across 50 OGBench tasks and 6 D4RL antmaze tasks. FAV achieves the best overall average performance, outperforming 9 competitive baselines across diverse robotics manipulation tasks. Despite being a single-step generator without an auxiliary generative model, FAV outperforms two baseline categories: multi-step generative policies (FAWAC, IFQL, QAM), which require iterative denoising, and distilled generative policies (SRPO, FQL), which are built on auxiliary generative models. Additionally, FAV surpasses DSRL, which additionally incurs inference-time overhead. The results show that FAV-Adaptive also surpasses all prior baselines on average. This makes FAV-Adaptive a strong default and a reliable starting point for hyperparameter search. The details of adaptive bandwidth configuration are elaborated in \Cref{app: automated bandwidth for FAV-adaptive}. For offline-to-online RL, \Cref{tab: offline_to_online_rl} demonstrates that FAV achieves the best post-adaptation performance, showing that the objective remains effective in online adaptation. 

We attribute the strong performance of FAV to three factors: FAV can utilize a powerful gradient signal of the Q-function as in reparameterized policy gradients~\citep{park2024value}; FAV framework enables a single-step policy, thereby avoiding the credit-assignment and backpropagation-through-time challenges of multi-step generative policies; and FAV does not rely on a complex auxiliary generative model or its distillation process that may harm the training stability and efficiency~\citep{dingconsistency,chenscore,fql_park2025}.

\paragraph{FAV as an attractive option for policy extraction.}
The results suggest that FAV is an attractive option for policy extraction, competitive with strong representative approaches such as reparameterized policy gradients \citep{park2024value, fql_park2025} and Adjoint Matching \citep{li2026q}. The comparison with FQL is particularly clean: FAV and FQL share the same codebase, model architectures, and single-step inference, differing only in the extraction objective. FQL relies on reparameterized policy gradients and an auxiliary flow-matching model for distillation; FAV directly amortizes sample-based variational inference. Despite this simpler mechanism, FAV outperforms FQL in both offline and offline-to-online RL, highlighting FAV as an effective class of policy extraction methods for RL.

\subsection{Conditional image generation}
\begin{figure}[t]
    \centering
    \includegraphics[width=\textwidth]{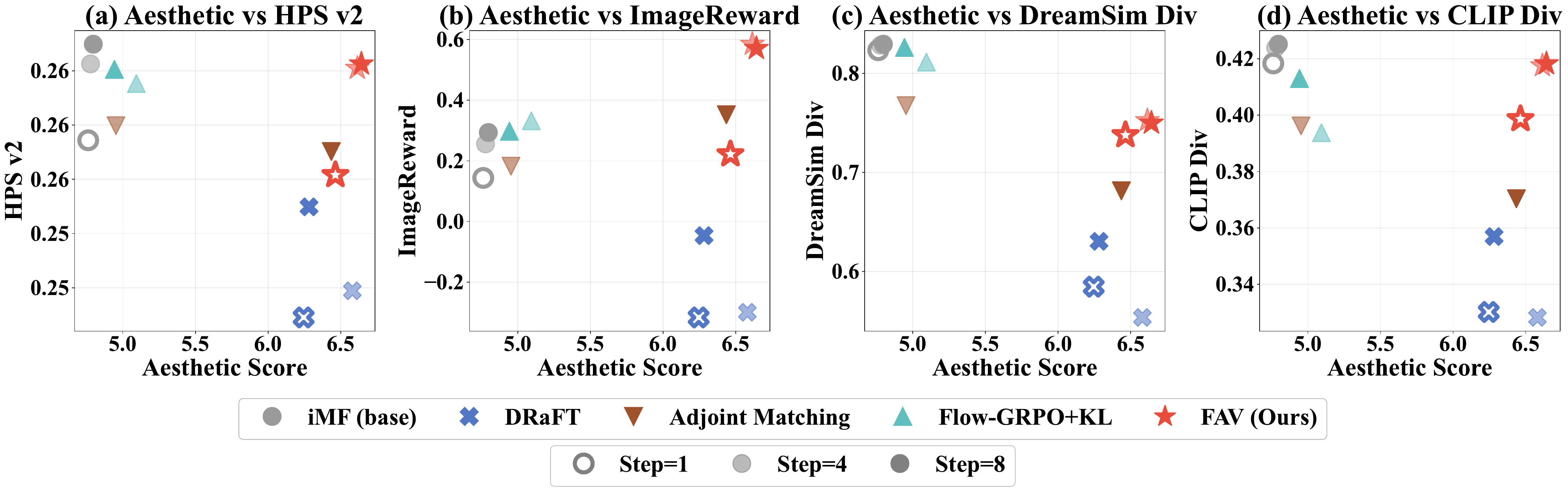}
    \caption{\textbf{Target reward vs evaluation metrics.} Aesthetic Score is the target reward; (a) HPSv2, (b) ImageReward, (c) DreamSim diversity and (d) CLIP diversity evaluate quality and diversity to indicate reward overoptimization. FAV achieves the best Pareto frontier.}
    \label{fig:imagenet_main}
    \vspace{-1em}
\end{figure}

\begin{table}[t]
\centering
\caption{\textbf{Comparison of inference-time alignment methods with FAV.} FAV can generate aligned samples with substantially lower inference cost than inference-time baselines.}
\resizebox{0.98\textwidth}{!}{
\begin{tabular}{lcccccc}
\toprule
Method & Time (s) $\downarrow$ & Aesthetic Score $\uparrow$ & HPSv2 $\uparrow$ & ImageReward $\uparrow$ & DreamSim Div $\uparrow$ & CLIP Div $\uparrow$\\
\midrule
BoN-256 & 28.6 & 5.68 (0.00) & \textbf{26.23 (0.00)} & 0.44 (0.00) & \textbf{0.82 (0.00)} & 0.41 (0.00)\\
ReNO-50 & 17.9 & 6.52 (0.01) & 26.01 (0.00) & 0.31 (0.03) & 0.80 (0.00) & \textbf{0.45 (0.00)} \\
\textbf{FAV (ours)} & \textbf{0.1} & \textbf{6.61 (0.09)} & 26.01 (0.00) & \textbf{0.58 (0.07)} & 0.75 (0.01) & 0.42 (0.01)\\
\bottomrule
\end{tabular}
}
\label{tab: imagenet_test_time}
\end{table}
Sampling from the reward-tilted distribution $p_{\text{ref}}(x)\exp(r(x))$ is a principled approach to image alignment, yielding high-reward samples while mitigating reward overoptimization~\citep{uehara2024understanding}. In this section, we demonstrate three key advantages of FAV: (1) superior alignment performance over existing baselines, (2) the preservation of fast generation speeds, and (3) generalizability across diverse model classes and reward settings. We provide implementation details in~\Cref{app: Implementation detail-Conditional Image Generation}.

\subsubsection{Experimental setup}
\paragraph{ImageNet-256} We fine-tune StyleGAN-XL \citep{sauer2022stylegan}, iMM \citep{zhouinductive}, iMeanFlow \citep{geng2025improved}, and drifting model \citep{deng2026generative} to optimize the LAION aesthetic score \citep{schuhmann2023laion}. To detect reward over-optimization, we measure the image quality using Human Preference Score (HPSv2) \citep{wu2023human} and ImageReward \citep{xu2023imagereward}, and diversity using mean pairwise cosine distance of DreamSim features (DreamSim Diversity) \citep{fu2023dreamsim}, and CLIP features (CLIP Diversity) \citep{radford2021learning}. We compare FAV against baseline alignment methods on iMeanFlow with step sizes of \{1, 4, 8\}.
% We conduct baseline comparisons on iMeanFlow for two reasons: it allows evaluation across varying step sizes {1,4,8}, and it is the only setting where existing alignment baselines can be approximately applied.

\paragraph{High-resolution text-to-image generation} We fine-tune Sana-sprint 1.6B \citep{chen2025sana} and evaluate on two distinct alignment tasks: (1) safety alignment using Falconsai NSFW image classifier \citep{falconsai2024nsfw} as the negative reward on SneakyPrompt \citep{yang2024sneakyprompt} and (2) human preference alignment, using HPSv2 as the target reward on DrawBench \citep{saharia2022photorealistic}. To indicate overoptimization, we employ ImageReward, Dreamsim Diversity, and CLIP Diversity. We conduct experiments with a step size of 4.

\paragraph{Baselines} We compare FAV against 4 categories of generative model alignment methods. \textbf{(1) Direct backpropagation}. DRaFT \citep{clarkdirectly} leverages reward gradients directly, which is efficient for few-step implicit models due to short gradient chains. \textbf{(2) RL-based methods}: Flow-GRPO \citep{liuflow} converts the deterministic Flow-ODE into a marginal-preserving SDE to measure transition likelihoods, enabling the use of the policy gradient method. RLCM \citep{oertell2024rl} applies policy gradient over the multistep sampling trajectory of consistency models \citep{song2023consistency, lusimplifying}. \textbf{(3) SoC-based methods}: Adjoint Matching \citep{domingoadjoint} converts the ODE into an SDE with memoryless noise schedules and uses the lean adjoint regression loss. \textbf{(4) Inference-time alignment methods}: Best-of-N \citep{karthik2023if} selects the highest-reward sample from N candidates, and ReNO \citep{eyring2024reno} optimizes the noise space using the reward gradient.

\paragraph{Evaluation} In the ImageNet-256 experiments, we randomly select 32 class labels from the full set of ImageNet classes and use them for both training and evaluation. In the text-to-image experiments, we randomly sample 32 prompts from SneakyPrompt and DrawBench to form the evaluation set, while training is conducted exclusively on the remaining prompts. For every metric, we generate 32 images for each of the 32 evaluation labels/prompts, and compute the metrics over the resulting 1,024 images. All experiments are conducted across 4 random seeds. 
% For DRaFT, to ensure a reliable comparison, we report the checkpoint immediately before reward overoptimization occurs.

\subsubsection{Results} 

\begin{wrapfigure}[19]{r}{0.50\textwidth}
    \vspace{-1.3em}
    \centering
    \small

    \captionof{table}{\textbf{FAV across diverse generator classes.} Evaluation metrics are reported at a comparable aesthetic score for a fair comparison.}
    \label{tab:imagenet_diverse_backbone}
    \resizebox{\linewidth}{!}{%
        % tables/imagenet_diverse_backbone.tex
\begin{tabular}{lccc}
\toprule
Model & Aesthetic $\uparrow$ & HPSv2 $\uparrow$ & Dreamsim $\uparrow$\\
\midrule
StyleGAN-XL(DRaFT) \cite{sauer2022stylegan} & 4.61 $\rightarrow$ 5.68 & 25.28 $\rightarrow$ 24.74 & 0.81 $\rightarrow$ 0.67 \\
StyleGAN-XL(FAV) \cite{sauer2022stylegan}   & 4.61 $\rightarrow$ \textbf{5.69} & 25.28 $\rightarrow$ \textbf{25.14} & 0.81 $\rightarrow$ \textbf{0.70} \\
\midrule
Drifting(DRaFT) \cite{deng2026generative}   & 4.71 $\rightarrow$ 6.20 & 25.36 $\rightarrow$ 24.37 & 0.82 $\rightarrow$ 0.56 \\
Drifting(FAV) \cite{deng2026generative}     & 4.71 $\rightarrow$ \textbf{6.27} & 25.36 $\rightarrow$ \textbf{24.89} & 0.82 $\rightarrow$ \textbf{0.68} \\
\midrule
iMM(DRaFT) \cite{zhouinductive}             & 4.73 $\rightarrow$ 6.17 & 25.67 $\rightarrow$ 24.98 & 0.83 $\rightarrow$ 0.67 \\
iMM(FAV) \cite{zhouinductive}               & 4.73 $\rightarrow$ \textbf{6.26} & 25.67 $\rightarrow$ \textbf{25.55} & 0.83 $\rightarrow$ \textbf{0.75} \\
\midrule
iMeanFlow(DRaFT) \cite{geng2025improved}    & 4.80 $\rightarrow$ 6.58 & 26.14 $\rightarrow$ 25.00 & 0.83 $\rightarrow$ 0.55 \\
iMeanFlow(FAV) \cite{geng2025improved}      & 4.80 $\rightarrow$ \textbf{6.61} & 26.14 $\rightarrow$ \textbf{26.01} & 0.83 $\rightarrow$ \textbf{0.75} \\
\bottomrule
\end{tabular}
\label{tab:imagenet_diverse_backbone}
    }

    \vspace{0.6em}

    \includegraphics[width=1\linewidth]{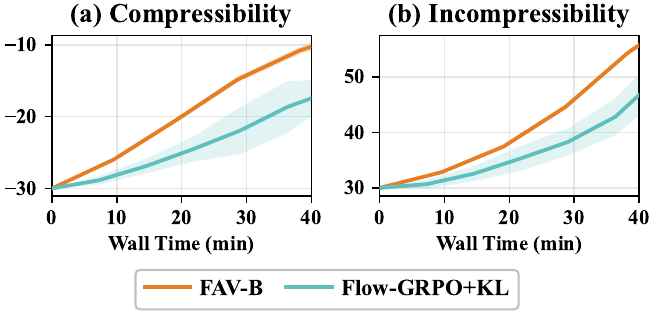}
    \captionof{figure}{\textbf{FAV-B on black-box reward functions.}}
    \label{fig:imagenet_non_differentiable_main}

    \vspace{-1em}
\end{wrapfigure}

\paragraph{FAV outperforms baselines in few-step regimes.}
\Cref{fig:imagenet_main} shows that FAV achieves the best Pareto frontier across \{1,4,8\} generation steps. DRaFT improves the target reward at all step sizes, but substantially degrades quality and diversity metrics. Adjoint Matching is competitive at 8-steps, but becomes less effective with fewer steps and is not applicable to the 1-step setting, consistent with \Cref{toy setting}. Notably, 1-step FAV already outperforms or matches 8-step baselines, demonstrating its effectiveness in the few-step regime.
Qualitative comparisons are provided in~\Cref{fig:imagenet_qualitative}.

\paragraph{FAV preserves fast generation.} As shown in~\Cref{tab: imagenet_test_time}, FAV generates samples approximately $180\times$ to $280\times$ faster than inference-time alignment methods with competitive image quality and diversity. By amortizing SVGD transport, FAV fully preserves the inference efficiency of few-step generative models.

\paragraph{FAV generalizes to diverse generators and reward functions.} We evaluate FAV on one-step noise-to-data mapping models (StyleGAN-XL, Drifting model), and 4-step flow map models (iMM, iMeanFlow). Among the fine-tuning baselines, DRaFT is the only method that can handle all model types, yet it cannot mitigate reward overoptimization. \Cref{tab:imagenet_diverse_backbone} shows that FAV consistently outperforms DRaFT in both target reward and evaluation metrics across all settings. Furthermore, as demonstrated in \Cref{fig:sana_main}, FAV outperforms the baselines on SANA-Sprint 1.6B alignment at $1024^2$ resolution, showing its scalability. We provide qualitative results in \Cref{fig:Sana_qualitative}.

\paragraph{FAV-B for black-box reward functions.}
We propose \textbf{FAV-B} to estimate the reward gradient with zeroth-order methods (See details in \Cref{App:FAV with non-differentiable reward}). As shown in \Cref{fig:imagenet_non_differentiable_main}, FAV-B effectively optimizes the compressibility and incompressibility rewards \cite{blacktraining} and outperforms Flow-GRPO+KL in aligning the 4-step iMeanFlow model, while achieving approximately $1.5\times$ faster training.

% \paragraph{Ablation and Sensitivity analysis} We provide the ablation study for each component of stein velocity field $\hat{\phi}_{q_\theta, q^*}^*(x)$ in \Cref{App: Ablation Analysis} and sensitivity analysis of $\beta$ and $\sigma$ in \Cref{App: Sensitivity Analysis}.

\section{Discussion}

\paragraph{Limitations.} FAV employs KDE-based nonparametric score estimation, whose consistency is guaranteed in the asymptotic regime. Hence, finite reference samples may introduce approximation error, particularly in high-dimensional domains. We mitigate this in image experiments by performing transport in a pretrained representation space, but this does not fully resolve the approximation error in high-dimensional data. Still, FAV consistently improves alignment across toy, generative policy, and image-generation settings, showing that the sample-based approximation is effective in practice.

\paragraph{Conclusion.} We introduce FAV, a novel sample-based alignment framework for few-step generative models. We approximate a reward-tilted distribution using SVGD with KDE, and amortize the resulting Stein transport into the generator to preserve fast generation. We evaluated FAV on two domains: generative policy alignment and image generator alignment. In robotics manipulation tasks, FAV outperformed multi-step flow policies and distilled policies on 56 offline RL tasks and 30 offline-to-online RL tasks, highlighting its effectiveness for policy extraction. In image alignment, FAV consistently achieved strong alignment performance while preserving sample quality and diversity across resolutions from 256$^2$ to 1024$^2$.

% \begin{ack}
% Use unnumbered first level headings for the acknowledgments. All acknowledgments

% Do {\bf not} include this section in the anonymized submission, only in the final paper. You can use the \texttt{ack} environment provided in the style file to automatically hide this section in the anonymized submission.
% \end{ack}

\clearpage

\bibliography{neurips_2026}
\bibliographystyle{plain}

\medskip

{
\small

%%%%%%%%%%%%%%%%%%%%%%%%%%%%%%%%%%%%%%%%%%%%%%%%%%%%%%%%%%%%

\clearpage
\section*{Appendix}
\appendix

\section{Implementation details}
\label[appendix]{}

In this section, we elaborate on implementation details for our experiments: toy setting, reinforcement learning, and image generator alignment.

\subsection{Toy experiments}
\label[appendix]{app: toy experiments}

\paragraph{Dataset.}
All experiments are conducted on the 8-Gaussians dataset \citep{lu2023contrastive}. The data distribution $p(x)$ is an equally weighted mixture of eight isotropic Gaussians whose centers are uniformly placed on a circle of radius $4$, with component standard deviation $0.5$. All samples are then globally rescaled by $1/\sqrt{2}$, yielding effective centers on a circle of radius $4/\sqrt{2} \approx 2.83$ and an effective per-component standard deviation of $0.5/\sqrt{2} \approx 0.354$. For training, we draw a pool of 500{,}000 samples once per run. For evaluation, we additionally draw a fixed reference set of 10{,}000 samples and reuse it for all metric computations during the run.

\paragraph{Reward function.}
The fine-tuning target is the reward-tilted distribution
\[
q^*(x) \propto p(x)\, \exp( r(x)),
\]
with $\beta = 1$. On the 8-Gaussians dataset, we use a soft cluster-assignment reward defined as
\[
\exp(r(x)) = \sum_{k=0}^{7} \mathrm{softmax}\!\left(-\|x-c_k\|^2\right)_k \cdot (k/7),
\]
where $\{c_k\}_{k=0}^{7}$ are the eight cluster centers defining the data distribution. Thus, cluster $k$ receives target reward $k/7$, making the eighth mode the most desirable and the first mode neutral. The softmax formulation makes the reward smooth and differentiable everywhere, although its gradient saturates near the cluster centers.

\paragraph{Pre-training.}
We pre-train VAE, Drifting model and MeanFlow on a fixed 2D toy dataset. All three models share a common backbone: 3-layer MLPs with hidden width 256 and SiLU  activations \citep{elfwing2018sigmoid}. They are optimized for 1M gradient steps with a batch size of $8192$.

\begin{itemize}
\item \textbf{Variational Autoencoder \citep{kingma2014auto}.} The VAE consists of an encoder $q_{\phi}(z|x)=\mathcal{N}(\mu_{\phi}(x), \mathrm{diag}(\exp\sigma_{\phi}^{2}(x)))$ and a decoder $p_{\theta}(x|z)$, both parameterized as three-layer SiLU MLPs with 256 hidden units. The latent dimension is $d_z=8$, and the prior is $p(z)=\mathcal{N}(0,I)$. We optimize the negative ELBO:
\[
\mathcal{L}_{\mathrm{VAE}}(x)=\frac{1}{2}\|x-\hat{x}\|_{2}^{2}+\beta\,\mathrm{KL}\!\left(q_{\phi}(z\mid x)\,\|\,\mathcal{N}(0,I)\right).
\]
We clamp $\mathrm{logvar}$ to $[-8,8]$ for numerical stability and use the reparameterization $z=\mu+\exp(\frac{1}{2}\sigma^{2})\epsilon$, where $\epsilon\sim\mathcal{N}(0,I)$. We set $\beta=0.02$.

\item \textbf{Drifting model \citep{deng2026generative}.}
We instantiate the drifting model as a one-step pushforward map $f_{\theta}:\mathbb{R}^{32}\to\mathbb{R}^{2}$ with latent noise
$\epsilon\sim\mathcal{N}(0,I_{32})$, parameterized by a three-layer SiLU MLP. Let $p_{\theta}=(f_{\theta})_{\#}p_{\epsilon}$ denote the generated distribution. Given positive samples $y^{+}\sim p_{\mathrm{data}}$ and negative samples
$y^{-}\sim p_{\theta}$, the drifting field is defined as
\[
V_{p_{\mathrm{data}},p_{\theta}}(x)
=
\frac{1}{Z_{p_{\mathrm{data}}}(x)Z_{p_{\theta}}(x)}
\mathbb{E}_{y^{+}\sim p_{\mathrm{data}},\,y^{-}\sim p_{\theta}}
\left[
k(x,y^{+})k(x,y^{-})(y^{+}-y^{-})
\right],
\]
where$
Z_{p_{\mathrm{data}}}(x)
:=
\mathbb{E}_{y^{+}\sim p_{\mathrm{data}}}
\left[k(x,y^{+})\right],~
Z_{p_{\theta}}(x)
:=
\mathbb{E}_{y^{-}\sim p_{\theta}}
\left[k(x,y^{-})\right].$
We use the kernel $k(x,y)=\exp(-\|x-y\|_2/\tau)$. The generator is trained by regression to the
stop-gradient drifted target:
\[
\mathcal{L}_{\mathrm{drift}}
=
\left\|
\operatorname{stopgrad}\!\left(
f_{\theta}(\epsilon)
+
V_{p_{\mathrm{data}},p_{\theta}}(f_{\theta}(\epsilon))-f_{\theta}(\epsilon)
\right)
\right\|_2^2 .
\]
We set the drifting temperature to $\tau=0.15$.

\item \textbf{MeanFlow \citep{gengmean}.} The MeanFlow model learns the mean velocity $u_{\theta}(x_t,r,t)$ of a linear interpolation coupling $x_t=(1-t)x_0+tx_1$ between data $x_0\sim p_{\mathrm{data}}$ at $t=0$ and noise $x_1\sim\mathcal{N}(0,I)$ at $t=1$. It is instantiated as a three-layer SiLU MLP taking $[x_t,t,t-r]$ as input. The pairs $(t,r)$ are drawn as $(\max,\min)$ of two $\mathrm{sigmoid}(\mathcal{N}(-0.4,1))$ samples, with probability $p_{\mathrm{eq}}=0.5$ replaced by $r=t$. Let $v_t=x_1-x_0$ denote the instantaneous velocity. The Jacobian-Vector Product (JVP) $\partial u_{\theta}/\partial t$ along $(v_t,0,1)$ gives the MeanFlow target
\[
u^{\star}(x_t,r,t)=\text{stopgrad}\!\left(v_t-(t-r)\frac{d u_{\theta}}{dt}\right).
\]
The objective is the adaptively weighted MSE:
\[
\mathcal{L}_{\mathrm{MF}}=\frac{\|u_{\theta}(x_t,r,t)-u^{\star}\|_{2}^{2}}{\left(\mathrm{stopgrad}\!\left(\|u_{\theta}-u^{\star}\|_{2}^{2}\right)+10^{-2}\right)^{p}},
\]
with adaptive-weighting power $p=1.0$. Gradients are clipped to $\|\cdot\|_2\leq 1$. 
\end{itemize}

\begin{table}[t]
\centering
\label{app: hyperparameters for toy experiments}
\caption{\textbf{Hyperparameters for toy experiments.}}
\begin{tabular}{lccc}
\toprule
\textbf{Component} & \textbf{VAE} & \textbf{Drifting model} & \textbf{MeanFlow} \\
\midrule
Optimizer & Adam & Adam & AdamW \\
$(\beta_1,\beta_2)$ & $(0.9,0.999)$ & $(0.9,0.999)$ & $(0.9,0.95)$ \\
Weight decay & 0 & 0 & 0 \\
Learning rate & $1\times 10^{-3}$ & $1\times 10^{-3}$ & $3\times 10^{-4}$ \\
Batch size & 8192 & 8192 & 8192 \\
Training steps & $1\times 10^{6}$ & $1\times 10^{6}$ & $1\times 10^{6}$ \\
Hidden width / depth & $256\ /\ 3$ & $256\ /\ 3$ & $256\ /\ 3$ \\
Activation & SiLU & SiLU & SiLU \\
Latent / input dim & $d_z=8$ & $d_z=32$ & --- \\
Gradient clipping & --- & --- & $\|g\|_2 \le 1$ \\

\bottomrule
\end{tabular}
\end{table}

\paragraph{Fine-tuning baselines.} For MeanFlow fine-tuning with Regularized REINFORCE and Adjoint Matching, we approximate the MeanFlow mapping $u_\theta(x_t,r,t)$ as an instantaneous velocity $v_t$, allowing us to formulate the sampling as an SDE. We inject Gaussian noise where the noise scale follows a VP-SDE-style \citep{songscore} schedule $ dx = -v(x,t)\,dt + \sigma(t)\,dW, ~~\sigma(t) = \eta\sqrt{{t}/{(1-t)}}$ for the regularized REINFORCE and memoryless schedule $\sigma(t) = \sqrt{{2t}/{(1-t)}}$ for the Adjoint Matching. Since applying Regularized REINFORCE and Adjoint Matching to one-step noise-to-data mapping is non-trivial, we evaluate the fine-tuning baselines only on the multi-step MeanFlow backbone. We fine-tune each algorithm for 20,000 gradient steps with a 1024 batch size and select the best checkpoint according to the lowest exponentially averaged KLD.

\begin{itemize}
    \item \textbf{Regularized REINFORCE.} 
    For Regularized REINFORCE, we implement the on-policy variant of Relative Trajectory Balance (RTB)~\citep{venkatraman2024amortizing}, which is equivalent to KL-regularized REINFORCE~\citep{deleu2025relative}. We use VarGrad~\citep{richter2020vargrad} to estimate the normalizing constant of the RTB objective. The noise scale $\eta$ is selected by grid search over $\{0.01, 0.05, 0.1, 0.2, 0.3, 0.5\}$. The selected values are $\eta=0.01$ for $n=2$, $\eta=0.05$ for $n=4$, $\eta=0.10$ for $n=8$, and $\eta=0.20$ for $n=16$, where $n$ denotes the number of iterative denoising steps.

    \item \textbf{Adjoint Matching.} In the case of Adjoint Matching, we follow the implementation guideline provided in the original paper \citep{domingoadjoint}. Specifically, we implement the memoryless noise schedule $\sigma(t) = \sqrt{{2(t+h)}/{(1-t+h)}}$, where $h$ denotes the time discretization step size, for numerical stability and faster fine-tuning. For gradient evaluation, we subsample half of the timesteps: we uniformly sample timesteps from the first $75\%$ of the trajectory and always include all timesteps in the final $25\%$. (See details in Appendix G in \citep{domingoadjoint}.)

\end{itemize}

\subsection{Offline and Offline-to-Online RL}
\label[appendix]{app: offline and offline-to-online RL}

We implement FAV on the codebases of FQL \footnote{https://github.com/seohongpark/fql} and QAM \footnote{https://github.com/ColinQiyangLi/qam} using the JAX \citep{jax2018github} implementation. We greatly thank the authors of \citep{fql_park2025} for providing highly reproducible codebases.  Please note that FAV shares the same architecture for the actor and critic, latent parameterization, optimizer hyperparameters, critic settings, and normalization technique, as well as the offline-to-online adaptation protocol, with FQL \citep{fql_park2025}.

\paragraph{Benchmarks and training protocol.}
We evaluate FAV on 56 offline RL tasks from OGBench and D4RL AntMaze. From OGBench, we use the reward-based single-task variants, covering 50 tasks across 10 environments, and additionally include 6 challenging AntMaze tasks from D4RL. For offline-to-online RL, we select the six OGBench environments on which FAV attains the lowest offline RL performance, corresponding to 30 tasks in total. Following the prior experimental evaluation protocol \citep{fql_park2025}, agents are trained for 1M gradient steps in the offline setting, and for offline-to-online RL we continue training for an additional 1M online steps. For D4RL AntMaze, agents are trained for 500K steps. We report results over 8 random seeds as the mean with 95\% confidence intervals.

\paragraph{Policy and critic architectures.}
We instantiate FAV as a single-step noise-to-action policy that maps the concatenation of the observation and latent noise directly to an action. The actor is implemented as a multilayer perceptron with 4 hidden layers of width 512, GELU activations, and no normalization layers. The critic is a double-Q ensemble with two parallel value heads, each using 4 hidden layers of width 512 and GELU activations, with LayerNorm applied after each activation. Both actor and critic use variance-scaling initialization with fan-average uniform kernels. The critic output layer is linear without a final activation, while the actor output is linear and clipped to the valid action range.

\paragraph{Latent parameterization and optimization.}
For action generation, we sample latent noise as $z\in\mathbb R^a \sim \mathcal{N}(0, I)$ and feed the concatenated vector $[s; z] \in \mathbb{R}^{s + a}$ to the actor, where $s$ and $a$ denote the dimension of state and action, respectively. Both the actor and the critic are optimized using Adam with a learning rate of $3 \times 10^{-4}$ and a batch size of 256. A target critic is updated at every gradient step using Polyak averaging with coefficient $\tau = 0.005$.

\paragraph{Critic training and normalization.}
We use a double-Q critic and compute the Bellman target using the mean of the target ensemble by default, although we also support a minimum aggregation variant. The target is given by
\[
y = r + \gamma (1-d)\,\mathrm{mean}\bigl(Q_{\mathrm{target}}(s', a')\bigr),
\]
where $a' \sim \pi(s')$, and $d$ denotes the termination flag.  The critic is trained with a mean-squared Bellman error. We use a discount factor $\gamma = 0.99$ by default and set $\gamma = 0.995$ for \texttt{antmaze-giant}, \texttt{humanoidmaze}, and \texttt{antsoccer} environments. We do not normalize observations, and actions are clipped to $[-1,1]$ during generation. Rewards are left unscaled except for D4RL AntMaze, where we apply the standard shift $r \leftarrow r - 1$. 

\paragraph{Offline-to-Online fine-tuning.}
For online adaptation, FAV uses the empirical action distribution induced by the replay buffer as the reference distribution and preserves the same core objective used in offline training. We use a single circular replay buffer initialized with the offline dataset and sample uniformly from it thereafter. The replay buffer capacity is 2M transitions, and we use an update-to-data ratio of 1, i.e., one gradient step per environment interaction. During online fine-tuning, exploration is induced solely through the stochastic one-step policy by sampling fresh Gaussian noise at each interaction step, without any additional exploration bonus or $\epsilon$-greedy strategy.

\paragraph{FAV hyperparameters.}
FAV introduces $\beta$ that controls the strength of value-guided alignment, along with a kernel bandwidth $\tau$. In our implementation, we sweep $\beta$ over \{0.5, 1, 2, 3, 5\}. We search the kernel bandwidth $\tau$ from \{0.05, 0.1, 0.5, 1.0\}. For FAV-Adaptive, we determine the kernel bandwidth directly from data using Scott's rule.

\paragraph{Kernel specification.}
FAV uses a Gaussian RBF kernel,
\begin{align}
\label{eq: FAV kernel}
k(x,y) = \exp\!\left(-\frac{\|x-y\|^2}{\tau}\right), \qquad \tau = 2\sigma^2,
\end{align}
to construct the Stein velocity field. We employ equal bandwidth for both of two kernels. For each state, we generate $N=8$ action particles and compute the drift using three components: a prior-score term, a value-gradient term, and a repulsive term between generated particles.

\begin{table*}[ht]                                                                                                                                                   
  \centering                                                                                      
  \small                                                                                                                                                              
  \setlength{\tabcolsep}{4pt}                                                                                                                                       
  \caption{\textbf{RL tasks hyperparameters} Methods   
  unavailable in the source are left as `--`.}                                                                                                                 
  \resizebox{\linewidth}{!}{%                                                                                                                                         
\begin{tabular}{lccccccccccc}                                                                                                                                       
  \toprule[1pt]                                                                                   
  & \multicolumn{2}{c}{Gaussian Policies}                                                                                                                             
  & \multicolumn{1}{c}{Latent Opt.}                                                                                                                                   
  & \multicolumn{3}{c}{Flow-based Policies}                                                       
  & \multicolumn{3}{c}{Distillation Policies}                                                                                                                         
  & \multicolumn{2}{c}{Ours} \\                                                    
  \cmidrule(lr){2-3} \cmidrule(lr){4-4} \cmidrule(lr){5-7} \cmidrule(lr){8-10} \cmidrule(lr){11-12}                                                                   
  Task                                                                                                                                                                
  & IQL & ReBRAC & DSRL & FAWAC & IFQL & QAM & SRPO & CAC & FQL & FAV-Adaptive & FAV \\                                                                                  
  & $\alpha$ & $(\alpha_1,\alpha_2)$ & $\sigma_z$ & $\alpha$ & $N$ & $\tau$ & $\beta$ & $\eta$ & $\alpha$ & $\beta$ & $(\tau,\beta)$  \\
\end{tabular}%                                                                                                                                                      
  }                                                                                                                                                                   
  \label{tab:transcribed_hparam_reordered}                                                        
  \end{table*}

\paragraph{Evaluation protocol.} We report results over 8 random seeds as mean $\pm$ 95\% confidence interval. For OGBench, agents are trained for 1M offline steps, and in the offline-to-online setting, we continue training for an additional 1M online steps. For D4RL antmaze, agents are trained for 500K steps. Following the protocol of Park et al. \citep{fql_park2025}, we report offline OGBench performance averaged over 800K, 900K, and 1M steps, and D4RL antmaze performance averaged over checkpoints at 300K, 400K, and 500K steps. For offline-to-online RL, we report performance at 1M steps and 2M steps. The results of baselines are carried over from prior works \citep{wagenmakersteering}, while we reproduce the QAM with the same critic and actor architecture and hyperparameters shared with FQL and FAV.

\paragraph{Baselines for Offline RL tasks.}
The results in \Cref{table:main_offline_rl} are drawn from multiple sources. Results for IQL, ReBRAC, FAWAC, IFQL, SRPO, and CAC are taken from Park et al.~\citep{fql_park2025}, which conducted an extensive hyperparameter search. DSRL results on OGBench are taken from Wagenmaker et al.~\citep{wagenmakersteering}.
Since the original DSRL paper~\citep{wagenmakersteering} does not report results on the D4RL antmaze tasks, we reproduce them using the QAM codebase~\citep{li2026q}. For consistency with our benchmark, we replace the 10-head Q-ensemble and Q-chunking~\citep{li2025decoupled} on the implementation of DSRL on the QAM codebase with a 2-head Q-function without Q-chunking. Following the guidelines of Li et al.~\citep{li2026q}, we grid-search $\sigma_z$ over \{0.1,0.2,0.4,0.6,0.8,1,1.2,1.4\}. We also reproduce QAM as a recent baseline. For a fair comparison, we again adopt a 2-head Q-function without Q-chunking, keeping the same policy and critic architectures as in the other baselines and FAV. Following~\citep{li2026q}, we grid-search $\tau$ over \{0.1, 0.3, 1, 3, 10, 30\}.

\paragraph{Baselines for Offline-to-Online RL tasks.} For the offline-to-online setting, we reproduce the full benchmark, including IQL, ReBRAC, RLPD, IFQL, and FQL—to obtain performance curves over training steps. Hyperparameters are shared between the offline and offline-to-online phases. All experiments are built on the official codebase of FQL.

\paragraph{Automated bandwidth for FAV-Adaptive}\label{app: automated bandwidth for FAV-adaptive}

FAV-Adaptive sets the temperature from the dataset statistics. We use Scott's rule with the action scale computed once from the full offline dataset $\mathcal{D}$. Let $d_a$ be the action dimension and 
$
\hat{\sigma}_{\mathcal{D}}
=
\frac{1}{d_a}
\sum_{j=1}^{d_a}
\operatorname{Std}_{a \sim \mathcal{D}}(a_j).
$
Given batch size $n$, we set
$
h_{\mathrm{Scott}}
=
n^{-1/(d_a+4)} \hat{\sigma}_{\mathcal{D}},
\tau_{\mathrm{adaptive}}
=
2h_{\mathrm{Scott}}^2.
$
FAV-Adaptive uses the full-dataset action statistics to determine a fixed environment-specific kernel temperature, whereas the sample-size factor scales with the training batch size.

\subsection{Conditional image generation}
\label[appendix]{app: Implementation detail-Conditional Image Generation}
\paragraph{Model specification} 
The following table lists the official implementations of the pre-trained models used in our experiments. For ImageNet experiments, we use the ImageNet 256$\times$256 pre-trained checkpoints provided in each repository.
\begin{center}
\label{tab:model_specification}
\begin{tabular}{ll}
\toprule
\textbf{Models} & \textbf{Links} \\
\midrule
StyleGAN-XL~\cite{sauer2022stylegan} & \url{https://github.com/autonomousvision/stylegan-xl} \\
Drifting Model-L/latent~\cite{deng2026generative} & \url{https://github.com/lambertae/drifting} \\
IMM-XL/2~\cite{zhouinductive} & \url{https://github.com/lumalabs/imm} \\
iMeanFlow-XL/2~\cite{geng2025improved} & \url{https://github.com/Lyy-iiis/imeanflow/tree/torch} \\
Sana-Sprint 1.6B~\cite{chen2025sana} & \url{https://github.com/NVlabs/Sana} \\
\bottomrule
\end{tabular}
\end{center}

\paragraph{Pre-trained encoder}
We instantiate the pre-trained encoder $\psi$ in~\Cref{eq:deep kernel FAV loss} with the vision backbone of each reward model, ensuring that kernel proximity reflects reward-relevant semantic similarity of samples. Specifically: (i) OpenAI CLIP ViT-L/14 for the LAION aesthetic reward \citep{schuhmann2023laion}; (ii) OpenCLIP ViT-H/14 (LAION-2B pre-trained) fine-tuned on Human Preference Dataset v2 for HPS \citep{wu2023human}; (iii) ViT-base (ImageNet-21k pre-trained) fine-tuned on $\sim$80K normal/NSFW images by Falconsai for NSFW \citep{falconsai2024nsfw}.

\paragraph{Batch for training} For each of $N_c$ conditions, we draw $N_\text{gen}$ samples from the current model $q_{\theta}$  and $N_\text{ref}$ samples from the pre-trained model $p_{\text{ref}}$. We compute the loss over these $N_\text{gen} + N_\text{ref}$ samples per condition and aggregate across $N_c$ conditions, yielding an effective batch size of $N_c \cdot (N_\text{gen} + N_\text{ref})$. In all experiments, we set $N_\text{gen} = N_\text{ref} = 16$ and $N_c = 8$, resulting in an effective batch size of $256$. All baseline methods are trained with the same effective batch size.

\paragraph{Hyperparameters for training} 
Table~\ref{tab:training_hparams} summarizes the training hyperparameters used in each experimental setting. When sampling from the pre-trained models, we follow the default hyperparameters (e.g. CFG scale) provided in each official repository. 
% For FAV, we use $N_\text{pos}, N_\text{gen} = 16$ and $N_c = 8$, yielding an effective batch size of $256$; all baseline methods are trained with the same effective batch size.
\begin{center}
\captionof{table}{Training hyperparameters for each experimental setting.}
\label{tab:training_hparams}
\resizebox{\linewidth}{!}{% tables/training_hparams.tex
\begin{tabular}{lccc}
\toprule
 & \textbf{ImageNet-Aesthetic} & \textbf{Sana-NSFW} & \textbf{Sana-HPS} \\
\midrule
Learning rate & 0.0005 & 0.0005 & 0.0005 \\
Optimizer & AdamW $(\beta_1=0.9,\beta_2=0.95)$ & AdamW $(\beta_1=0.9,\beta_2=0.95)$ & AdamW $(\beta_1=0.9,\beta_2=0.95)$ \\
Weight decay & 0.01 & 0.01 & 0.01 \\
Gradient clip norm & 2 & 2 & 1 \\
Max steps & 200 & 100 & 200 \\
Batch sizes & 256 & 256 & 256 \\
\bottomrule
\end{tabular}}
\end{center}

\paragraph{Hyperparameters for FAV}
Two key hyperparameters in FAV are the alignment coefficient $\beta$, which controls the strength of reward guidance, and the Gaussian RBF kernel bandwidth $\tau$ in \Cref{eq: FAV kernel}, which determines the locality of SVGD interactions. We search over $\beta =\{0.01, 0.05, 0.1, 0.5, 1, 5, 10, 50, 100\}$ and $\tau = \{0.01, 0.1, 0.3, 0.5, 1\}$. We find that $\tau = 0.3$ empirically works well across all settings, and therefore fix it throughout our experiments.  For LoRA fine-tuning \citep{hu2022lora}, we select the rank and scale parameters based on the model backbone and the target reward. Table~\ref{tab:image_FAV_hyperparameter} reports the selected values for each experiment.

\begin{center}
\captionof{table}{Hyperparameters for FAV in conditional image generation}
\label{tab:image_FAV_hyperparameter}
\resizebox{1\textwidth}{!}{%
  % tables/FAV_hyperparameter.tex
\begin{tabular}{lcccccccc}
\toprule
 & \multicolumn{4}{c}{\textbf{Aesthetic}} & \textbf{Compressibility} & \textbf{Incompressibility} & \textbf{NSFW} & \textbf{HPS} \\
\cmidrule(lr){2-5} \cmidrule(lr){6-6} \cmidrule(lr){7-7} \cmidrule(lr){8-8} \cmidrule(lr){9-9}
Backbone & StyleGAN & Drifting & IMM & iMeanFlow & iMeanFlow & iMeanFlow & Sana-sprint & Sana-sprint \\
$\beta$ & 10 & 1 & 0.5 & 0.5 & 0.01 & 0.05 & 0.1 & 100 \\
$\tau$ & 0.3 & 0.3 & 0.3 & 0.3 & 0.3 & 0.3 & 0.3 & 0.3 \\
LoRA rank & 32 & 32 & 16 & 16 & 16 & 16 & 8 & 32 \\
LoRA scale & 1 & 4 & 4 & 4 & 4 & 4 & 1 & 1 \\
\bottomrule
\end{tabular}
}
\end{center}

\paragraph{Baselines for conditional image generation tasks} We compare FAV against two families of alignment methods: fine-tuning approaches (DRaFT, Flow-GRPO+KL, RLCM, and Adjoint Matching) and test-time search approaches (Best-of-N and ReNO). 
% For fine-tuning methods, we sweep over the key hyperparameters of each method to report their best results; for test-time search methods, we directly follow the official implementations, since they do not modify the pre-trained model.

\begin{itemize}
    \item \textbf{DRaFT}~\citep{clarkdirectly}: When reward overoptimization occurs too rapidly, we multiply the reward by a scaling factor to ensure a fair comparison with other baselines. 
    
    \item \textbf{Flow-GRPO+KL}~\citep{liuflow}: To approximate the sampling procedure of the flow-map model as ODE solving, we treat iMeanFlow's average velocity field as the instantaneous velocity field follwing \Cref{app: toy experiments}. We search over the noise level  $\eta = \{0.01, 0.05, 0.1, 0.2\}$ and $\beta = \{1e-6, 1e-5, 1e-4, 5e-4, 1e-3, 1e-2, 0.5, 1.0\} $. We set the group size as 32.
    
    \item \textbf{RLCM}~\citep{oertell2024rl}: We adopt the RLCM MDP, replacing the consistency model~\citep{song2023consistency} based policy with its sCM~\citep{lusimplifying} counterpart, since our pre-trained model Sana-Sprint~\citep{chen2025sana} is built on the sCM framework. All other components follow the official implementation.
    
    \item \textbf{Adjoint Matching}~\citep{domingoadjoint}: We fine-tune the flow-map model using the same approximation as Flow-GRPO+KL. We adopt the practical memoryless noise schedule, timestep selection to prioritize the last 25\% of the sampling trajectory, and apply loss clipping via threshold following Appendix G in~\citep{domingoadjoint}. We search over LCT scalar $= \{1.6, 16, 160\}$ and $\beta = \{1, 5, 10, 50, 100, 500, 1000\}$.
    
    \item \textbf{Best-of-N}: We generate $N = 256$ samples per prompt from the pre-trained model and select the one with the highest reward.
    
    \item \textbf{ReNO}~\citep{eyring2024reno}: We follow the official implementation with 50 search steps, a learning rate of 5.0, and employ the Nesterov momentum optimizer \citep{nesterov2013introductory, sutskever2013importance} with a regularizer coefficient of 0.01.
\end{itemize}

\paragraph{Prompts} 
We report the class labels and prompts used in our experiments. For ImageNet256 experiments with aesthetic score as the target reward, we use 32 randomly sampled ImageNet class labels for both training and evaluation, motivated by the "simple animals" setting commonly used in diffusion alignment~\citep{blacktraining, clarkdirectly}. The class labels reported in \Cref{tab:imagenet_labels}.

For high-resolution text-to-image experiments, we consider two target rewards.
When optimizing HPS, we construct an evaluation set by sampling 32 prompts from the DrawBench prompts \citep{saharia2022photorealistic}, with the number of prompts sampled from each category proportional to its category size. The sampled DrawBench prompts are listed in \Cref{tab:drawbench_prompts}.
When optimizing the NSFW classifier, we use SneakyPrompt \citep{yang2024sneakyprompt}.
We do not disclose the exact prompts, as they are adversarially constructed to elicit unsafe generations. We randomly sample 32 prompts for evaluation. In both tasks, training and evaluation prompts are disjoint.

\begin{table}[h]
\centering
\caption{\textbf{Class labels used for ImageNet256 experiments.}}
\label{tab:imagenet_labels}
\footnotesize
\begin{tabular}{r l r l r l r l}
\toprule
\textbf{ID} & \textbf{Class} &
\textbf{ID} & \textbf{Class} &
\textbf{ID} & \textbf{Class} &
\textbf{ID} & \textbf{Class} \\
\midrule
0   & tench             & 9   & ostrich             & 22  & bald eagle          & 39  & common iguana \\
55  & green snake       & 69  & trilobite           & 80  & black grouse        & 105 & koala \\
108 & sea anemone       & 115 & sea slug            & 130 & flamingo            & 207 & golden retriever \\
291 & lion              & 387 & lesser panda        & 398 & abacus              & 403 & aircraft carrier \\
404 & airliner          & 409 & analog clock        & 414 & backpack            & 483 & castle \\
497 & church            & 540 & drilling platform   & 547 & electric locomotive & 550 & espresso maker \\
561 & forklift          & 562 & fountain            & 620 & laptop              & 649 & megalith \\
650 & microphone        & 671 & mountain bike       & 732 & Polaroid camera     & 985 & daisy \\
\bottomrule
\end{tabular}
\end{table}
\begin{longtable}{p{0.05\linewidth} p{0.88\linewidth}}
\caption{\textbf{DrawBench prompts used for high-resolution text-to-image generation.}}
\label{tab:drawbench_prompts}\\
\toprule
\# & Prompt \\
\midrule
\endfirsthead

\toprule
\# & Prompt \\
\midrule
\endhead

1 & A blue bird and a brown bear. \\
2 & A pink colored giraffe. \\
3 & A white colored sandwich. \\
4 & A yellow book and a red vase. \\
5 & Rainbow coloured penguin. \\
6 & An elephant under the sea. \\
7 & Two cats and two dogs sitting on the grass. \\
8 & Two cats and one dog sitting on the grass. \\
9 & Two cats and three dogs sitting on the grass. \\
10 & A triangular purple flower pot. A purple flower pot in the shape of a triangle. \\
11 & A side view of an owl sitting in a field. \\
12 & A cube made of brick. A cube with the texture of brick. \\
13 & An instrument used for cutting cloth, paper, and other thin material, consisting of two blades laid one on top of the other and fastened in the middle so as to allow them to be opened and closed by a thumb and finger inserted through rings on the end of their handles. \\
14 & An organ of soft nervous tissue contained in the skull of vertebrates, functioning as the coordinating center of sensation and intellectual and nervous activity. \\
15 & A long curved fruit which grows in clusters and has soft pulpy flesh and yellow skin when ripe. \\
16 & Paying for a quarter-sized pizza with a pizza-sized quarter. \\
17 & A donkey and an octopus are playing a game. The donkey is holding a rope on one end, the octopus is holding onto the other. The donkey holds the rope in its mouth. A cat is jumping over the rope. \\
18 & Tcennis rpacket. \\
19 & Pafrking metr. \\
20 & A giraffe underneath a microwave. \\
21 & A banana on the left of an apple. \\
22 & A carrot on the left of a broccoli. \\
23 & Jentacular. \\
24 & Painting of the orange cat Otto von Garfield, Count of Bismarck-Schönhausen, Duke of Lauenburg, Minister-President of Prussia. Depicted wearing a Prussian Pickelhaube and eating his favorite meal---lasagna. \\
25 & Illustration of a mouse using a mushroom as an umbrella. \\
26 & Greek statue of a man tripping over a cat. \\
27 & Darth Vader playing with raccoon in Mars during sunset. \\
28 & McDonalds Church. \\
29 & A realistic photo of a Pomeranian dressed up like a 1980s professional wrestler with neon green and neon orange face paint and bright green wrestling tights with bright orange boots. \\
30 & A storefront with ``Diffusion'' written on it. \\
31 & A storefront with ``Google Research Pizza Cafe'' written on it. \\
32 & A storefront with ``Hello World'' written on it. \\

\bottomrule
\end{longtable}

\clearpage

\section{KDE-based score estimation}
\label[appendix]{App: Score estimation through Gaussian kernel}
We write the KDE-based score approximation in population form to emphasize the resulting Stein velocity field in \Cref{KDE for Intractable Score Estimation}. In practice, the expectation over $p_{\rm ref}$ is replaced by an empirical average over finite reference samples. Accordingly, this section derives the KDE-based score estimator in its finite-sample form.
\subsection{Derivation of KDE-based score estimator}
Let $p$ be a density on $\mathbb{R}^d$ and let $k_\sigma$ denote the Gaussian kernel with bandwidth $\sigma > 0$:
\begin{align}
k_\sigma(u) = (2\pi\sigma^2)^{-d/2}\exp\!\left(-\frac{\|u\|^2}{2\sigma^2}\right).  
\end{align}
Convolving $p$ with $k_\sigma$ gives the smoothed density:
\begin{align}\label{eq: smoothed}
p_\sigma(x) = (k_\sigma * p)(x) = \int k_\sigma(x - u)\,p(u)\,du.
\end{align}
Given i.i.d.\ samples $\{X_i\}_{i=1}^N$ from $p$, the \textbf{kernel density estimator} (KDE) approximates $p_\sigma$ by replacing the expectation with an empirical average:
\begin{align}\label{eq: kde}
\hat{p}_\sigma(x) = \frac{1}{N}\sum_{i=1}^N k_\sigma(x - X_i),
\end{align}
so that $p_\sigma(x) = \mathbb{E}[\hat{p}_\sigma(x)]$. We now derive the KDE-based score estimator for $\nabla_x \log p(x)$, the score of the true density $p$. For the Gaussian kernel, differentiation gives:
\begin{align}\label{eq: nabla gaussian kernel}
\nabla_x k_\sigma(x - y) = \frac{y - x}{\sigma^2}\,k_\sigma(x - y).
\end{align}
Differentiating $\hat{p}_\sigma$ and dividing by $\hat{p}_\sigma(x)$ yields the score estimate:
\begin{align}
\nabla_x \log \hat{p}_\sigma(x)
= \frac{\nabla_x \hat{p}_\sigma(x)}{\hat{p}_\sigma(x)} 
= \sum_{i=1}^N
\frac{k_\sigma(x - X_i)}
{\sum_j k_\sigma(x - X_j)}
\frac{X_i - x}{\sigma^2}
= \sum_{i=1}^N
\hat{\tilde{k}}_\sigma(x, X_i)
\frac{X_i - x}{\sigma^2},
\end{align}
which corresponds to the classical mean-shift vector \citep{cheng1995mean}. Here $\hat{\tilde{k}}_\sigma$ are normalized weights summing to~1, making this a weighted average of directions $(X_i - x)$ with nearby points weighted more heavily.

In the FAV setting, the density of interest is the reference distribution $p_{\rm ref}$, and the score is evaluated at a model sample $x' \sim q_\theta$. Thus, using reference samples $x_i^{\rm ref}\sim p_{\rm ref}$, the reference score can be approximated as
\begin{align}
\nabla_{x'}\log p_{\rm ref}(x')
\approx
\sum_{i=1}^N
\hat{\tilde k}_\sigma(x',x_i^{\rm ref})
\frac{x_i^{\rm ref}-x'}{\sigma^2},
\end{align}
where the empirical normalized weight is defined as
\begin{align}
\hat{\tilde k}_\sigma(x',x_i^{\rm ref})
:=
\frac{k_\sigma(x'-x_i^{\rm ref})}
{\sum_{j=1}^N k_\sigma(x'-x_j^{\rm ref})}.
\end{align}
Equivalently, in expectation notation, this becomes
\begin{align}
\label{eq: KDE-based score estimation}
\nabla_{x'}\log p_{\rm ref}(x')
\approx
\mathbb E_{x^{\rm ref}\sim p_{\rm ref}}
\left[
\tilde k_\sigma(x',x^{\rm ref})
\frac{x^{\rm ref}-x'}{\sigma^2}
\right],
\end{align}
with the population-level normalized weight
\begin{align}
\tilde k_\sigma(x',x^{\rm ref})
:=
\frac{k_\sigma(x'-x^{\rm ref})}
{\mathbb E_{\bar x^{\rm ref}\sim p_{\rm ref}}
\left[k_\sigma(x'-\bar x^{\rm ref})\right]}.
\end{align}
The empirical weight $\hat{\tilde k}_\sigma$ is the finite-sample estimator of the population-level weight $\tilde k_\sigma$, and converges to $\tilde k_\sigma$ as $N \to \infty$.
Substituting KDE-based approximation in~\Cref{eq: KDE-based score estimation} into the prior-alignment term of \Cref{eq: optimal transport vector for reward tilted distribution} recovers the form used in \Cref{eq: approximated stein velocity field}.

\subsection{Consistency condition of KDE-based surrogate target distribution}
\label[appendix]{app: Consistency condition of KDE-based surrogate target distribution}
Recall that the true target distribution is $q^*(x)\propto p_{\text{ref}}(x)\exp(\beta r(x))$ and KDE-based surrogate target distribution is $\hat q_\sigma^*(x)\propto \hat p_\sigma(x)\exp(\beta r(x))$. In this section, we establish that the $\hat q_\sigma^*$ consistently recovers the $q^*$ under standard KDE regularity conditions.

For notational simplicity, we write $p$ for $p_{\mathrm{ref}}$ and define
\begin{align}
w(x) := \exp(\beta r(x)).
\end{align}
Then the true and surrogate target distributions are given by
\begin{align}
q^*(x)
=
\frac{p(x)w(x)}{Z},
\qquad
Z :=
\int p(x)w(x)\,dx,
\end{align}
and
\begin{align}
\hat q_\sigma^*(x)
=
\frac{\hat p_\sigma(x)w(x)}{\hat Z_\sigma},
\qquad
\hat Z_\sigma :=
\int \hat p_\sigma(x)w(x)\,dx.
\end{align}

We assume that (i) $p$ is a valid density on $\mathbb{R}^d$, (ii) the kernel
$K$ is sufficiently smooth with finite moments; in particular, the Gaussian
kernel $k_\sigma$ satisfies these conditions, (iii) the reward weight is bounded,
i.e., $\|w\|_\infty < \infty$, and (iv) the normalizing constant is finite and
strictly positive, i.e., $0<Z<\infty$.

Under these conditions, the KDE-induced tilted distribution is consistent for
the true tilted distribution in total variation distance:
\begin{align}
\|\hat q_\sigma^* - q^*\|_1
\xrightarrow{p} 0,
\qquad
\text{as } \sigma \to 0 \text{ and } N\sigma^d \to \infty .
\label{eq:tilted_distribution_consistency}
\end{align}

\noindent\textit{\textbf{Proof.}}\;
By standard $L^1$-consistency results for kernel density estimators, we have
\begin{align}
\|\hat p_\sigma - p\|_1
=
\int |\hat p_\sigma(x)-p(x)|\,dx
\xrightarrow{p} 0,
\label{eq:kde_l1_consistency}
\end{align}
as $\sigma\to 0$ and $N\sigma^d\to\infty$
\citep{devroye1985nonparametric, vandermeulen2013consistency}.

We first show that the normalizing constant of the surrogate target distribution converges to that of the true target distribution. By
\eqref{eq:kde_l1_consistency} and the boundedness of $w$, we have
\begin{align}
|\hat Z_\sigma - Z|
&=
\left|
\int
\left(\hat p_\sigma(x)-p(x)\right)w(x)\,dx
\right| \\
&\le
\|w\|_\infty
\int |\hat p_\sigma(x)-p(x)|\,dx \\
&=
\|w\|_\infty
\|\hat p_\sigma-p\|_1
\xrightarrow{p} 0 .
\end{align}
Therefore,
\begin{align}
\hat Z_\sigma \xrightarrow{p} Z .
\label{eq:normalizer_consistency}
\end{align}

We now bound the $L^1$ distance between the surrogate and true tilted
distributions:
\begin{align}
\|\hat q_\sigma^* - q^*\|_1
&=
\int
\left|
\frac{\hat p_\sigma(x)w(x)}{\hat Z_\sigma}
-
\frac{p(x)w(x)}{Z}
\right|dx \\
&\le
\int
\left|
\frac{\hat p_\sigma(x)w(x)}{\hat Z_\sigma}
-
\frac{p(x)w(x)}{\hat Z_\sigma}
\right|dx
+
\int
\left|
\frac{p(x)w(x)}{\hat Z_\sigma}
-
\frac{p(x)w(x)}{Z}
\right|dx \\
&=
\frac{1}{\hat Z_\sigma}
\int w(x)|\hat p_\sigma(x)-p(x)|\,dx
+
\left|
\frac{1}{\hat Z_\sigma}
-
\frac{1}{Z}
\right|
\int p(x)w(x)\,dx \\
&\le
\frac{\|w\|_\infty}{\hat Z_\sigma}
\|\hat p_\sigma-p\|_1
+
Z
\left|
\frac{1}{\hat Z_\sigma}
-
\frac{1}{Z}
\right| .
\label{eq:tilted_l1_bound}
\end{align}
By \eqref{eq:normalizer_consistency}, $\hat Z_\sigma \xrightarrow{p} Z>0$.
Hence, $1/\hat Z_\sigma = O_p(1)$. Together with
\eqref{eq:kde_l1_consistency}, the first term in
\eqref{eq:tilted_l1_bound} converges to zero in probability. The second term
also converges to zero by \eqref{eq:normalizer_consistency} and the continuous
mapping theorem applied to the map $z\mapsto 1/z$, which is continuous at
$Z>0$. Therefore,
\begin{align}
\|\hat q_\sigma^* - q^*\|_1
\xrightarrow{p} 0 .
\end{align}

Since the total variation distance satisfies
$d_{\mathrm{TV}}(\hat q_\sigma^*,q^*)=\frac{1}{2}\|\hat q_\sigma^*-q^*\|_1$, this proves that the KDE-induced surrogate target distribution $\hat q^*_\sigma$ consistently recovers the true reward-tilted target distribution $q^*$ in total variation distance.

\clearpage

\section{FAV for black-box rewards}
\label[appendix]{App:FAV with non-differentiable reward}
In this section, we describe how FAV can be extended to black-box rewards. Recall that the optimal Stein velocity field:
\begin{align}
\label{eq: optimal Stein velocity field appendix}
\phi^*_{q_\theta, q^*}(x) 
&= \mathbb{E}_{x' \sim q_\theta} \Big[ \underbrace{k(x', x) \nabla_{x'} \log p_{\text{ref}}(x')}_{\text{Prior Alignment}} + \underbrace{\beta \cdot k(x', x) \nabla_{x'} r(x')}_{\text{Reward Guidance}} + \underbrace{\nabla_{x'} k(x', x)}_{\text{Diversity Enforcement}} \Big]
\end{align}
requires the first-order gradient $\nabla_{x'} r(x')$ of the reward function. In the case of a black-box reward function, we replace the exact gradient with a zeroth-order estimator based on Gaussian smoothing. Specifically, given a smoothing scale $\eta > 0$, we approximate the gradient of the Gaussian-smoothed reward
$\hat r_\eta(x') := \mathbb{E}_{\epsilon \sim \mathcal{N}(0, I)}[r(x' + \eta\epsilon)]$, 
which admits the closed-form expression:
\begin{equation}
    \nabla_{x'} \hat r_\eta(x') 
    = \frac{1}{\eta}\, \mathbb{E}_{\epsilon \sim \mathcal{N}(0, I)}
    [r(x' + \eta\epsilon)\,\epsilon].
\end{equation}
We estimate this expectation using a two-point zeroth-order estimator with symmetric perturbations \citep{nesterov2017random}:
\begin{equation}
\label{eq:zeroth-order gradient estimator}
    \nabla_{x'} \hat r_\eta(x') \approx \frac{1}{K}\sum_{k=1}^{K} \frac{r(x' + \eta u_k) - r(x' - \eta u_k)}{2\eta}\, u_k, \quad u_k \sim \mathcal{N}(0, I).
\end{equation}
This estimator only requires forward evaluations of $r$, making it applicable to arbitrary black-box rewards. We substitute this estimate $\nabla_{x'} \hat r_\eta(x')$ for $\nabla_{x'} r(x')$ and proceed with the standard FAV training objective.

In practice, the choice of perturbation space is important for reducing the variance of the zeroth-order estimate in high-dimensional data, such as images. Whenever possible, we apply the symmetric perturbations in the pre-trained representation space used by FAV. When perturbing the representation space is not feasible, we instead apply the perturbations in the latent space produced by the generative model $p_{\text{ref}}$ before VAE decoding. This avoids performing zeroth-order estimation in the raw pixel space, where the estimator can suffer from large variance.

To validate the effectiveness of our zeroth-order gradient estimator, we conduct experiments using iMeanFlow (4-step) on ImageNet256 with three target rewards: aesthetic score, compressibility, and incompressibility. For the aesthetic score, which is differentiable but can also be treated as a black-box reward, we compare three settings: (i) FAV (Ours), which uses the exact first-order gradient; (ii) FAV-B, which uses the zeroth-order estimator in Equation~\ref{eq:zeroth-order gradient estimator} with $\eta = 0.005$ and $K = 16$; and (iii) Flow-GRPO+KL, a representative black-box reward optimization baseline. In this setting, perturbations and FAV training are performed in the pre-trained CLIP space used by the aesthetic score. For compressibility and incompressibility, which are fully black-box rewards, we compare only FAV-B and Flow-GRPO+KL, and perform perturbations and training in the latent space prior to VAE decoding.

As shown in~\Cref{fig:imagenet_non_differentiable}, FAV-B improves the target reward using only forward evaluations of $r$. Although it is less efficient than first-order FAV when reward gradients are available, it achieves substantially higher reward than Flow-GRPO+KL for all settings within the same wall time, demonstrating FAV's versatility in black-box reward settings.

\begin{figure}[h]
    \centering
    \includegraphics[width=1\textwidth]{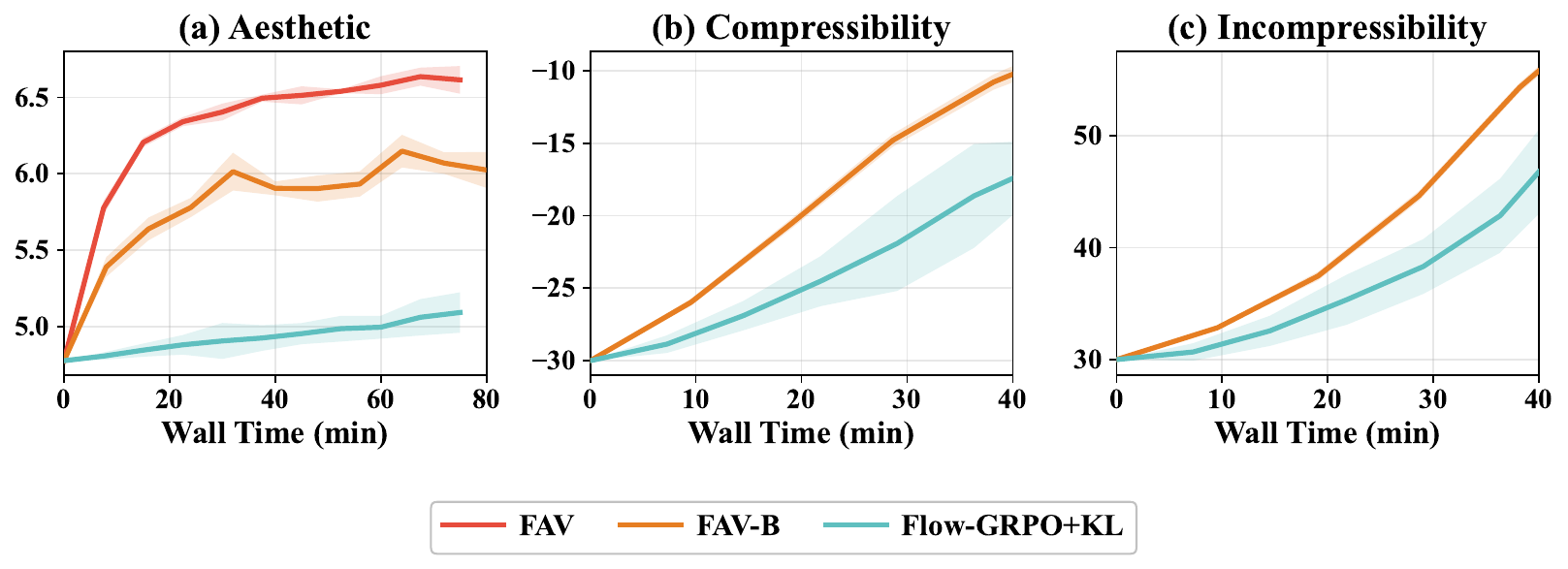}
    \caption{\textbf{FAV with non-differentiable reward.} (a) For aesthetic-score optimization, all methods are trained for 200 steps. (b),(c) For compressibility and incompressibility, FAV-B is trained for 40 steps, while Flow-GRPO+KL is run for the same wall-clock time, corresponding to 120 steps. Wall-clock time is measured on 4 NVIDIA RTX 3090 GPUs.}
    \label{fig:imagenet_non_differentiable}
\end{figure}

\clearpage

\section{Full result of RL tasks}
\label[appendix]{app: full result of rl tasks}
In this section, we present the full result tables for the offline RL and offline-to-online RL experiments. \Cref{tab:app_rl_full_results} reports the complete offline RL results across 50 OGBench tasks and 6 D4RL AntMaze tasks. DSRL results on OGBench are omitted because the original paper \citep{wagenmakersteering} does not provide per-task performance. \Cref{tab:app_off2on_final_full} reports the complete offline-to-online RL results, while \Cref{fig:app_offline_to_online_training_curve} presents the training curves of FAV and the baselines in the offline-to-online setting.

\begin{table*}[ht]                                                                                                                                                   
    \centering                                                                                       
    \small                                                                                                                                                            
    \setlength{\tabcolsep}{4pt}                                                                                                                                       
    \caption{\textbf{Offline RL performance.}}                                                                                                               
    \resizebox{\linewidth}{!}{%                                                                                                                                       
\begin{tabular}{lccccccccccc}                                                                                                                                     
    \toprule[1pt]                                                                                                                                                     
    & \multicolumn{2}{c}{Gaussian Policies}                                                          
    & \multicolumn{1}{c}{Latent Opt.}                                                                                                                                 
    & \multicolumn{3}{c}{Flow-based Policies}                                                                                                                         
    & \multicolumn{3}{c}{Distillation Policies}                                                                                                                       
    & \multicolumn{2}{c}{Ours} \\                                                                                                                                     
    \cmidrule(lr){2-3} \cmidrule(lr){4-4} \cmidrule(lr){5-7} \cmidrule(lr){8-10} \cmidrule(lr){11-12}                                                                 
    Task                                                                                                                                                              
    & IQL & ReBRAC & DSRL & FAWAC & IFQL & QAM & SRPO & CAC & FQL & FAV-Adaptive & FAV \\
\end{tabular}%                                                                                                                                                    
    }                                                                                                                                                                 
    \label{tab:app_rl_full_results}                                                                 
  \end{table*}    

\clearpage

\begin{table*}[ht]
\centering
\small
\setlength{\tabcolsep}{4pt}
\caption{\textbf{Offline-to-Online RL performance.} The left value denotes performance after 1M offline training steps, and the right value denotes performance after an additional 1M steps of online fine-tuning.}
\label{tab:app_off2on_final_full}
\resizebox{\linewidth}{!}{%
\begin{tabular}{lccccccc}
\toprule[1pt]
Task (offline 1M $\rightarrow$ online 1M) & IQL & ReBRAC & RLPD & IFQL & QAM & FQL & FAV \\
\midrule[1pt]
antmaze-giant-navigate-singletask-task1-v0  & 0 \std{0} $\rightarrow$ 0 \std{0} & 52 \std{16} $\rightarrow$ 99 \std{1} & 0 \std{0} $\rightarrow$ 14 \std{23} & 0 \std{0} $\rightarrow$ 0 \std{0} & 0 \std{0} $\rightarrow$ 0 \std{0} & 0 \std{1} $\rightarrow$ 30 \std{40} & 4 \std{6} $\rightarrow$ 68 \std{8} \\
antmaze-giant-navigate-singletask-task2-v0 & 0 \std{0} $\rightarrow$ 1 \std{2} & 15 \std{11} $\rightarrow$ 99 \std{1} & 0 \std{0} $\rightarrow$ 67 \std{26} & 0 \std{0} $\rightarrow$ 0 \std{0} & 1 \std{1} $\rightarrow$ 5 \std{3} & 0 \std{0} $\rightarrow$ 96 \std{2} & 29 \std{18} $\rightarrow$ 88 \std{5} \\
antmaze-giant-navigate-singletask-task3-v0 & 0 \std{0} $\rightarrow$ 0 \std{0} & 32 \std{32} $\rightarrow$ 98 \std{3} & 0 \std{0} $\rightarrow$ 81 \std{14} & 0 \std{0} $\rightarrow$ 0 \std{0} & 0 \std{0} $\rightarrow$ 0 \std{0} & 0 \std{0} $\rightarrow$ 6 \std{15} & 5 \std{5} $\rightarrow$ 34 \std{14} \\
antmaze-giant-navigate-singletask-task4-v0 & 0 \std{0} $\rightarrow$ 0 \std{1} & 2 \std{5} $\rightarrow$ 100 \std{0} & 0 \std{0} $\rightarrow$ 77 \std{30} & 0 \std{0} $\rightarrow$ 0 \std{0} & 0 \std{0} $\rightarrow$ 1 \std{1} & 0 \std{0} $\rightarrow$ 2 \std{4} & 30 \std{13} $\rightarrow$ 86 \std{3} \\
antmaze-giant-navigate-singletask-task5-v0 & 20 \std{7} $\rightarrow$ 13 \std{7} & 52 \std{15} $\rightarrow$ 99 \std{1} & 0 \std{0} $\rightarrow$ 82 \std{31} & 10 \std{9} $\rightarrow$ 0 \std{0} & 18 \std{4} $\rightarrow$ 48 \std{29} & 0 \std{0} $\rightarrow$ 85 \std{32} & 72 \std{10} $\rightarrow$ 96 \std{2} \\
\midrule
antsoccer-arena-navigate-singletask-task1-v0 & 17 \std{7} $\rightarrow$ 10 \std{4} & 0 \std{0} $\rightarrow$ 0 \std{0} & 0 \std{0} $\rightarrow$ 74 \std{8} & 65 \std{20} $\rightarrow$ 85 \std{5} & 93 \std{3} $\rightarrow$ 97 \std{3} & 78 \std{7} $\rightarrow$ 97 \std{2} & 84 \std{7} $\rightarrow$ 99 \std{1} \\
\end{tabular}%
}
\end{table*}

\clearpage

\begin{figure}[t]
    \centering
    \label{fig:app_offline_to_online_training_curve}
    \includegraphics[width=\textwidth]{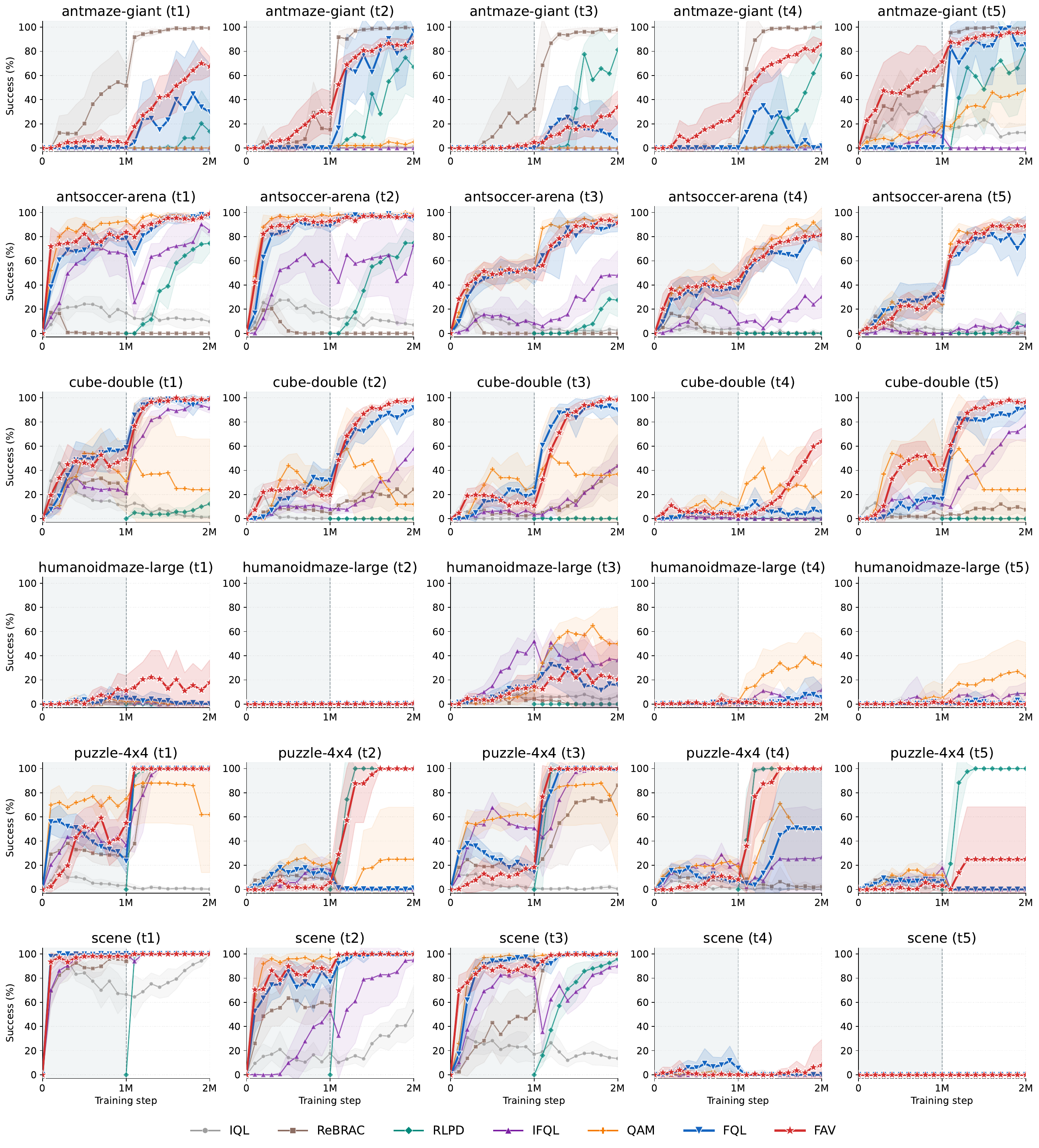}
    \caption{\textbf{Training curves for offline-to-online RL.}}
\end{figure}

\clearpage

\section{Computation analysis}
We analyze the computational cost of FAV by comparing its training and inference times with baselines on RL and image-generator alignment experiments.

\label[appendix]{app: computation analysis of offline rl}
\subsection{Offline RL}
We evaluate a total of eight offline RL methods, including FAV, on the \texttt{scene-play-single} environment. To ensure a fair comparison, all methods are executed under identical configurations on a single NVIDIA RTX 3090 GPU, with results averaged over 8 random seeds.

\begin{table}[ht]
    \centering
    \caption{Training and inference time for each method on {\texttt{scene-play-single}.}}
    \label{tab:computation_analysis}
    \smallskip
    \centering
    \begin{tabular}{lcc}
        \toprule
        \textbf{Method} & \textbf{Training Time (ms/step)} & \textbf{Inference Time (ms/step)} \\
        \midrule
        FQL    & 2.65 & 0.56 \\
        IFQL   & 1.96 & 0.85 \\
        IQL    & 2.08 & 0.60 \\
        ReBRAC & 1.51 & 0.59 \\
        RLPD   & 1.14 & 0.51 \\
        DSRL   & 1.96 & 0.64 \\
        QAM    & 4.67 & 0.78 \\
        \midrule
        FAV (Ours) & 3.05 & 0.53 \\
        \bottomrule
    \end{tabular}
\end{table}

\subsection{Image generator alignment}
We conduct a computational analysis on seven image-generator alignment algorithms, including FAV. For a fair comparison, all fine-tuning methods are evaluated with an effective batch size of 256 per step. Experiments using the iMeanFlow backbone are run on 4 NVIDIA RTX 3090 GPUs, while experiments using the SANA-Sprint backbone are run on 4 NVIDIA RTX 4090 GPUs.

\begin{table}[ht]
    \centering
    \caption{Training and inference time for each method.}
    \label{tab:computation_analysis}
    \smallskip
    \resizebox{\linewidth}{!}{%
    \begin{tabular}{lllcc}
        \toprule
        \textbf{Backbone} & \textbf{Method} & \textbf{Reward} & \textbf{Training Time (s/step)} & \textbf{Inference Time (s/step)} \\
        \midrule
        iMF (4 steps) & Flow-GRPO+KL     & Aesthetic       & 16.43 & 0.1 \\
        iMF (4 steps) & Flow-GRPO+KL     & Compressibility & 21.53 & 0.1 \\
        iMF (4 steps) & DRaFT            & Aesthetic       & 28.35 & 0.1 \\
        iMF (4 steps) & Adjoint Matching & Aesthetic       & 29.95 & 0.1 \\
        iMF (4 steps) & Best-of-256      & Aesthetic       & --    & 28.6 \\
        iMF (4 steps) & ReNO-50          & Aesthetic       & --    & 17.9 \\
        SANA-Sprint 1.6B (4 steps) & RLCM           & HPS             & 23.38    & 0.6 \\
        SANA-Sprint 1.6B (4 steps) & RLCM           & NSFW            & 23.14    & 0.6 \\
        SANA-Sprint 1.6B (4 steps) & DRaFT          & HPS             & 40.34    & 0.6 \\
        SANA-Sprint 1.6B (4 steps) & DRaFT          & NSFW            & 39.93    & 0.6 \\
        \midrule
        iMF (4 steps) & FAV (Ours)       & Aesthetic       & 20.12 & 0.1 \\
        iMF (4 steps) & FAV (Ours)       & Compressibility & 60.14 & 0.1 \\
        SANA-Sprint 1.6B (4 steps) & FAV (Ours)     & HPS             & 28.63 & 0.6 \\
        SANA-Sprint 1.6B (4 steps) & FAV (Ours)     & NSFW            & 28.34 & 0.6 \\

        \bottomrule
    \end{tabular}
    }
\end{table}

\clearpage

\section{Ablation analysis}
\label[appendix]{App: Ablation Analysis}
In this section, we analyze the effect of each component in the optimal transport vector of FAV: the prior alignment term, the reward guidance term, and the repulsive term in \Cref{eq: optimal transport vector for reward tilted distribution}. As shown in \Cref{fig:imagenet_ablation}, removing the prior alignment term ("w/o prior") leads to substantially higher reward scores but causes severe mode collapse. Without the reward guidance term ("w/o reward"), FAV fails to optimize the target reward, with the aesthetic score remaining flat throughout training. Removing the repulsive term ("w/o repulsive") yields a small but consistent decrease in diversity compared to the full model, suggesting that it offers a marginal contribution to diversity preservation on top of the prior alignment term. Experiments are conducted using iMeanFlow (4 steps) on ImageNet 256 with the aesthetic score as the target reward, averaged over 4 seeds.

\begin{figure}[h]
    \centering
    \includegraphics[width=0.8\textwidth]{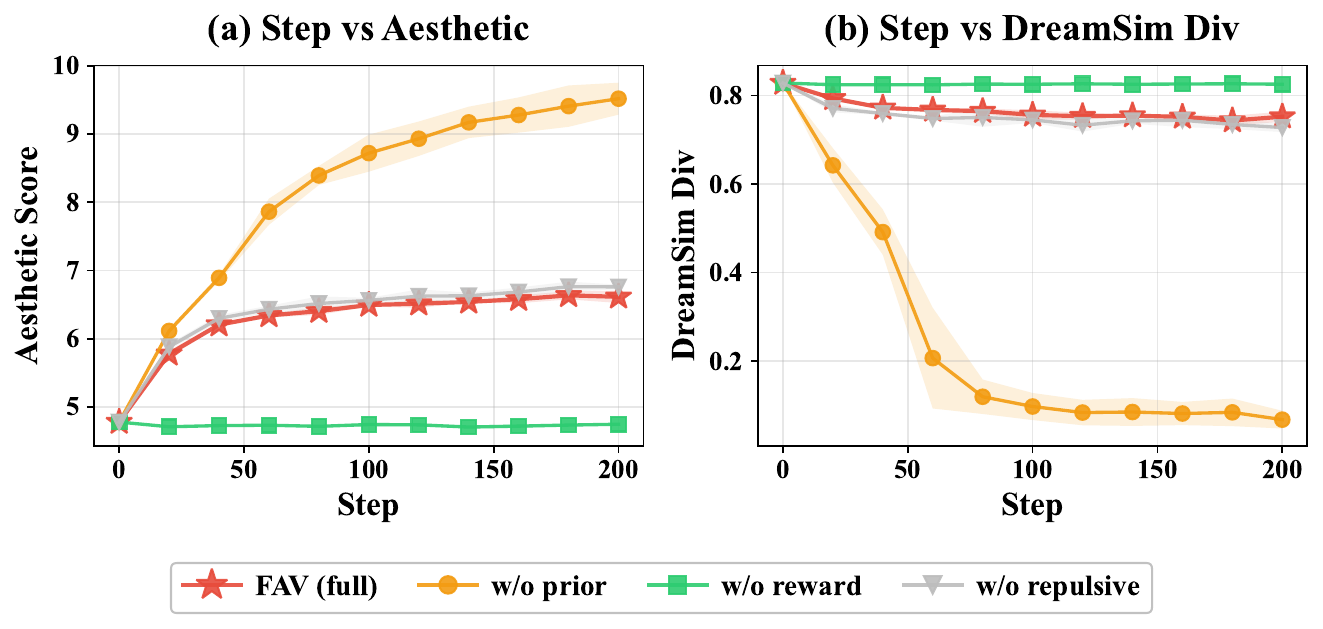}
    \caption{\textbf{Ablation analysis for each components of FAV.}}
    \label{fig:imagenet_ablation}
\end{figure}

\clearpage

\section{Sensitivity analysis}
\label[appendix]{App: Sensitivity Analysis}
In this section, we conduct a sensitivity analysis on the key hyperparameters of FAV: the temperature parameter $\beta$ and the kernel bandwidth $\tau$. Experiments are conducted using iMeanFlow (4 steps) on ImageNet 256 with the aesthetic score as the target reward, averaged over 4 seeds. As shown in \Cref{fig:imagenet_sensitivity_beta}, $\beta$ controls the reward alignment strength, enabling a tunable reward–diversity tradeoff: larger $\beta$ pushes samples more aggressively toward high-reward modes, while smaller $\beta$ stays closer to the prior.

The kernel bandwidth $\tau$, defined as $\tau := 2\sigma^2$, controls the interaction scale of the kernel $k_\sigma$. As shown in \Cref{fig:imagenet_sensitivity_tau}, increasing $\tau$ generally improves reward optimization but reduces sample diversity. We attribute this trade-off to the locality of kernel interactions. When $\tau$ is small, samples mainly interact with nearby reference points, preserving the multi-modal structure of the prior and maintaining diversity. However, such local updates limit reward improvement because samples are unlikely to move toward higher-reward modes outside their local neighborhoods. In contrast, when $\tau$ is large, the update becomes overly global, smoothing interactions across distinct modes. This weakens mode separation and makes samples more likely to collapse toward the single highest-reward mode, thereby reducing diversity.

\begin{figure}[h]
    \centering
    \includegraphics[width=0.8\textwidth]{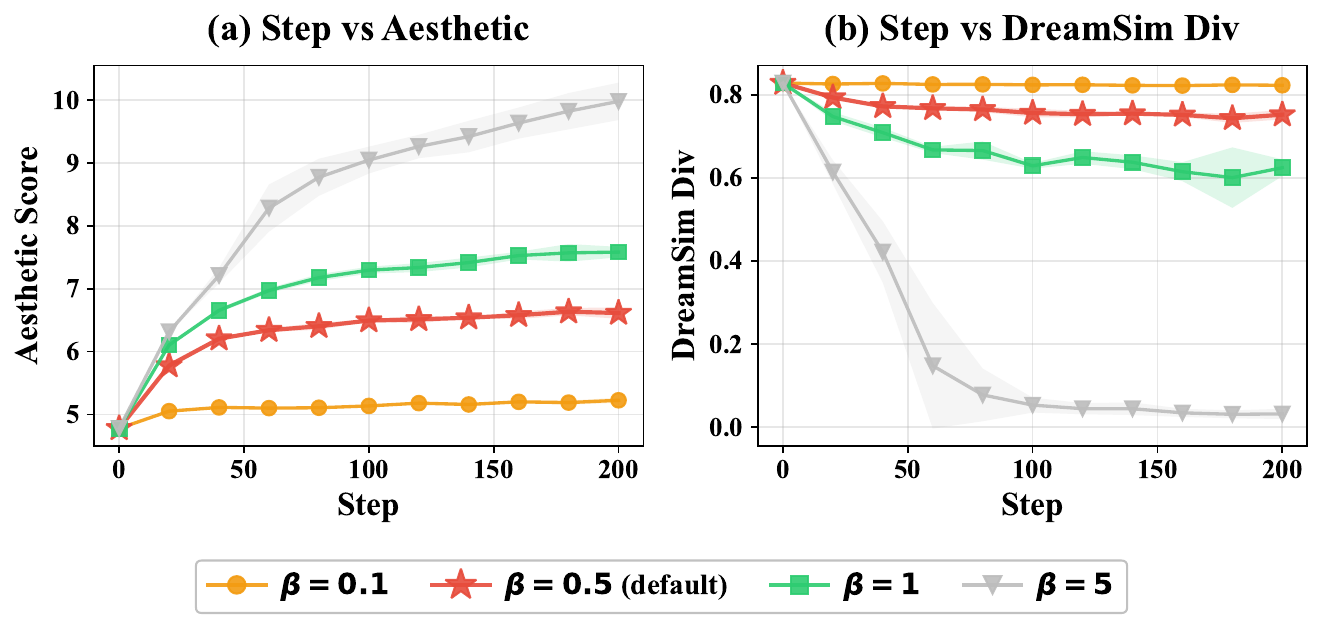}
    \caption{\textbf{Sensitivity analysis for temperature parameter $\beta$.}}
    \label{fig:imagenet_sensitivity_beta}
\end{figure}

\begin{figure}[h]
    \centering
    \includegraphics[width=0.8\textwidth]{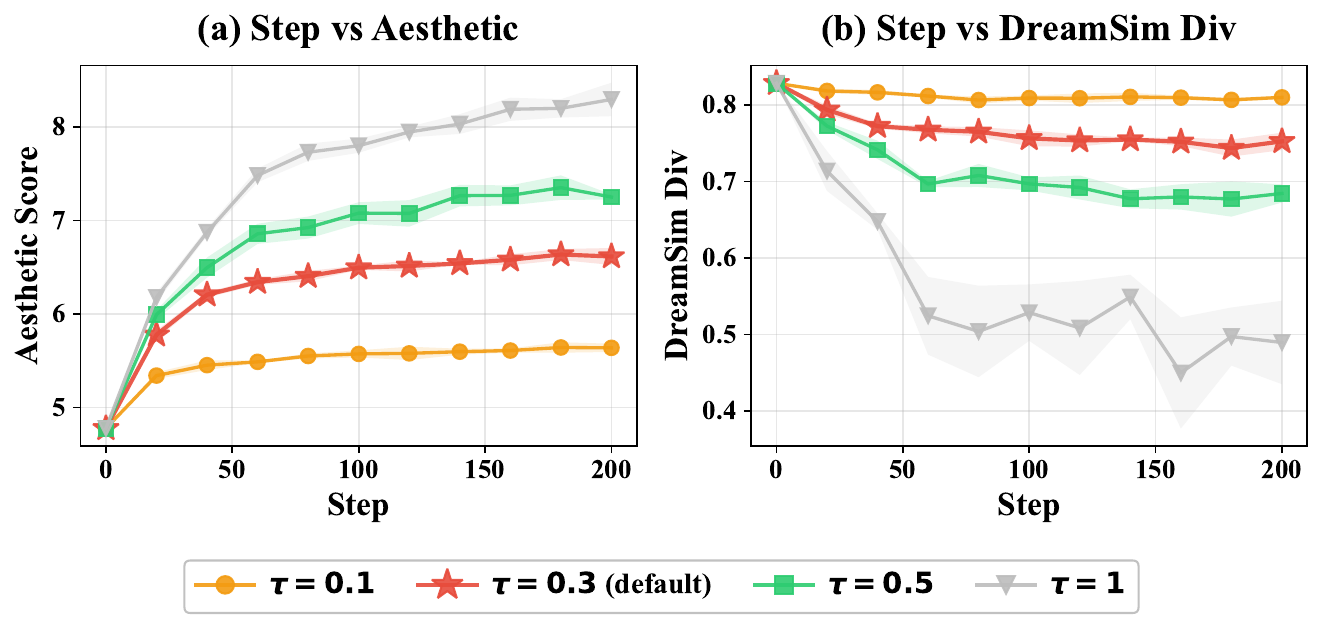}
    \caption{\textbf{Sensitivity analysis for kernel bandwidth $\tau$}.}
    \label{fig:imagenet_sensitivity_tau}
\end{figure}

We additionally evaluate FAV's sensitivity on the RL side using three environments: \texttt{puzzle-3x3-play} (5 tasks; center $\tau^\star{=}0.5,\,\beta^\star{=}0.5$), D4RL \texttt{antmaze-large-play} (1 task; $\tau^\star{=}1,\,\beta^\star{=}3$), and \texttt{antsoccer-arena-navigate} (5 tasks; $\tau^\star{=}1,\,\beta^\star{=}1$). For each environment, we perform two axis-aligned 1D sweeps about its center: one varies $\beta\in\{0.5,1,2,3\}$ with $\tau{=}\tau^\star$, the other varies $\tau\in\{0.05,0.1,0.5,1\}$ with $\beta{=}\beta^\star$. Each cell of \Cref{tab: rl sensitivity} reports mean $\pm$ std over 4 seeds at 1M offline gradient steps; centers are marked $^\star$ and per-column best in bold.

\begin{table}[h]
\centering
\small
\setlength{\tabcolsep}{4pt}
\caption{\textbf{RL hyperparameter sensitivity.}}
\label{tab: rl sensitivity}
\resizebox{\linewidth}{!}{%
\begin{tabular}{l ccc | l ccc}
\toprule
$\beta$ & {puzzle-3x3} & {antmaze-l} & {antsoccer} & $\tau$ & {puzzle-3x3} & {antmaze-l} & {antsoccer} \\
\midrule
$0.5$ & $\mathbf{73.4 \pm 23.4}^\star$ & $\phantom{0}2.0 \pm \phantom{0}2.8$ & $54.9 \pm 13.1$            & $0.05$ & $22.9 \pm 39.5$              & $\phantom{0}0.0 \pm \phantom{0}0.0$ & $\phantom{0}0.0 \pm \phantom{0}0.0$ \\
$1.0$ & $64.3 \pm 29.4$               & $30.0 \pm 11.2$                    & $\mathbf{60.0 \pm 24.3}^\star$ & $0.10$ & $21.3 \pm 39.5$              & $\phantom{0}0.0 \pm \phantom{0}0.0$ & $\phantom{0}0.0 \pm \phantom{0}0.0$ \\
$2.0$ & $27.6 \pm 31.0$               & $\mathbf{64.5 \pm \phantom{0}6.6}$ & $39.6 \pm 31.9$            & $0.50$ & $\mathbf{73.4 \pm 23.4}^\star$ & $\mathbf{62.5 \pm \phantom{0}3.4}$  & $54.4 \pm 12.2$ \\
$3.0$ & $14.7 \pm 22.5$               & $58.0 \pm 38.7^\star$              & $17.1 \pm 18.0$            & $1.00$ & $61.0 \pm 24.1$              & $58.0 \pm 38.7^\star$               & $\mathbf{60.0 \pm 24.3}^\star$ \\
\bottomrule
\end{tabular}%
}
\end{table}

The sensitivity study shows that both $\beta$ and $\tau$ are important hyperparameters for FAV. In particular, too small $\tau$ can substantially degrade performance, as very small values make the kernel nearly one-hot and lead to collapse. However, this concern is mitigated in practice by FAV-Adaptive, where $\tau$ is selected automatically using Scott's rule, as demonstrated in our main experiments. This leaves $\beta$, which controls the strength of the $Q$-gradient in the reward-tilted transport, as the primary hyperparameter to tune. We note that such a reward-guidance coefficient is not unique to FAV: analogous hyperparameters appear in most baselines \cite{dingconsistency, fql_park2025, li2026q}. Despite using a relatively small search range of $\beta$, FAV achieves strong performance across environments, suggesting that its sensitivity remains manageable in practice.

\clearpage

\section{High-resolution text-to-image alignment results}
In this section, we compare fine-tuning methods (DRaFT \citep{clarkdirectly}, RLCM \citep{oertell2024rl}) with FAV on high-resolution text-to-image generation. Experiments are conducted using Sana-Sprint 1.6B (4 steps) with the NSFW classifier and HPS as target rewards. As shown in \Cref{fig:sana_main}, FAV achieves the highest target reward while preserving the image quality and diversity.
\begin{figure}[h]
    \centering
    \includegraphics[width=\textwidth]{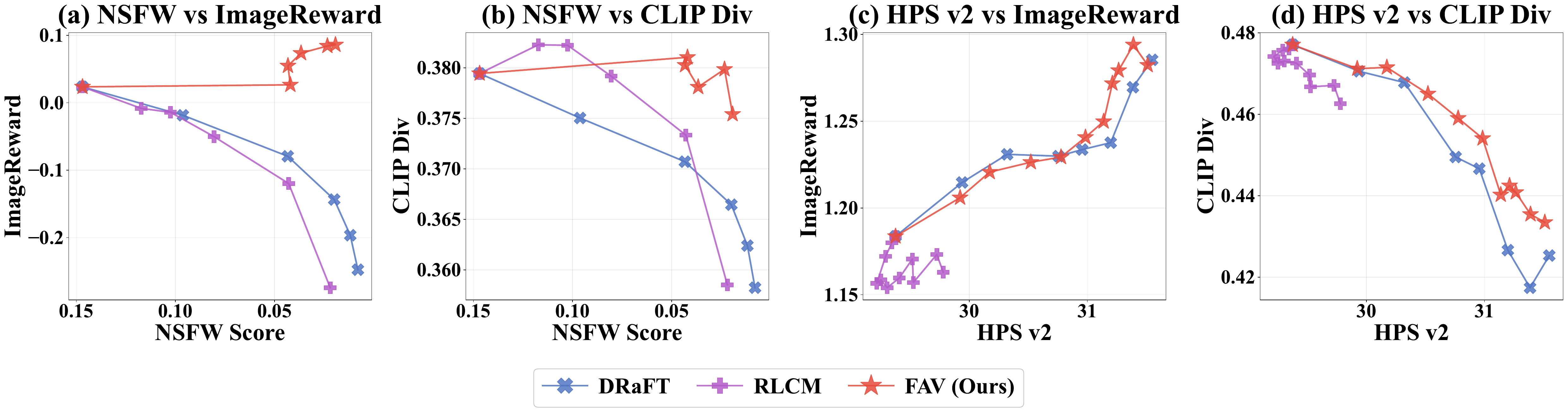}
    \caption{\textbf{Training dynamics of each alignment method.} NSFW classifier as the target reward for (a),(b); HPS as the target reward for (c),(d).}
    \label{fig:sana_main}
\end{figure}
\clearpage

\section{Qualitative analysis}
In this section, we provide qualitative results of the image generator alignment experiments. 
\subsection{ImageNet-256}
Figure~\ref{fig:imagenet_qualitative} shows qualitative comparisons on ImageNet 256, where each method aligns the iMeanFlow (8 steps) model with the aesthetic score as the target reward. 
\begin{figure}[h]
    \centering
    \vspace{0.8em}
    \includegraphics[width=1\linewidth]{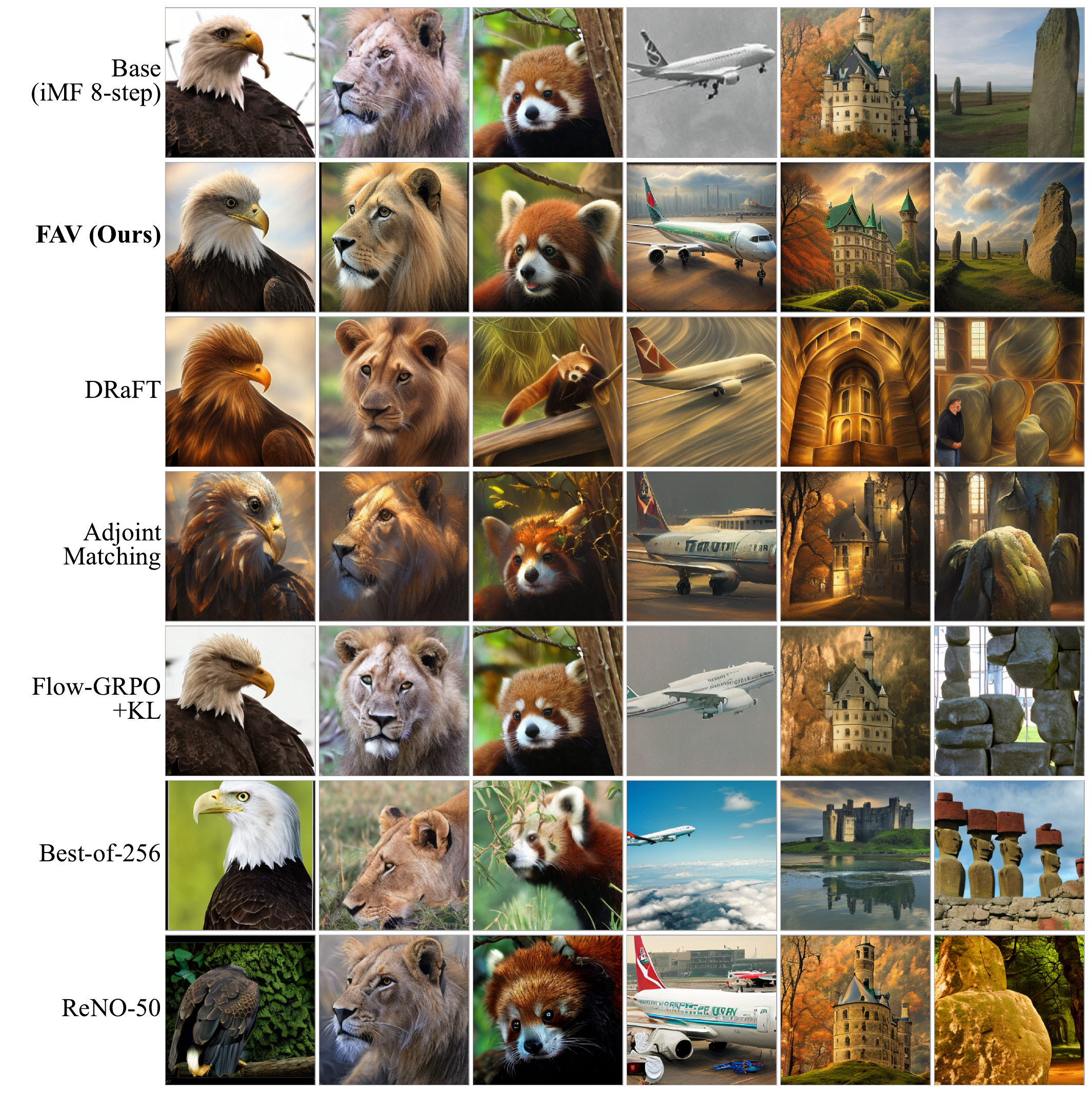}
    \vspace{-0.0em}
    \captionof{figure}{\textbf{Qualitative comparison on ImageNet 256}.}
    \label{fig:imagenet_qualitative}
\end{figure}
\clearpage

\subsection{High-resolution text-to-image generation}
Figure~\ref{fig:Sana_qualitative} shows qualitative samples from Sana-Sprint 1.6B (4-step) fine-tuned with each alignment method. The first four rows show samples from models fine-tuned on DrawBench prompts to maximize the HPS reward, and the last row shows samples from models fine-tuned on Sneaky prompts to minimize the NSFW classifier score. Since Sneaky prompts contain adversarial prompt and cannot be disclosed here, we refer readers to the official repository\footnote{\url{https://github.com/Yuchen413/text2image_safety}} for the prompt used in the last row.

\begin{figure}[ht]
    \centering
    \includegraphics[width=\textwidth]{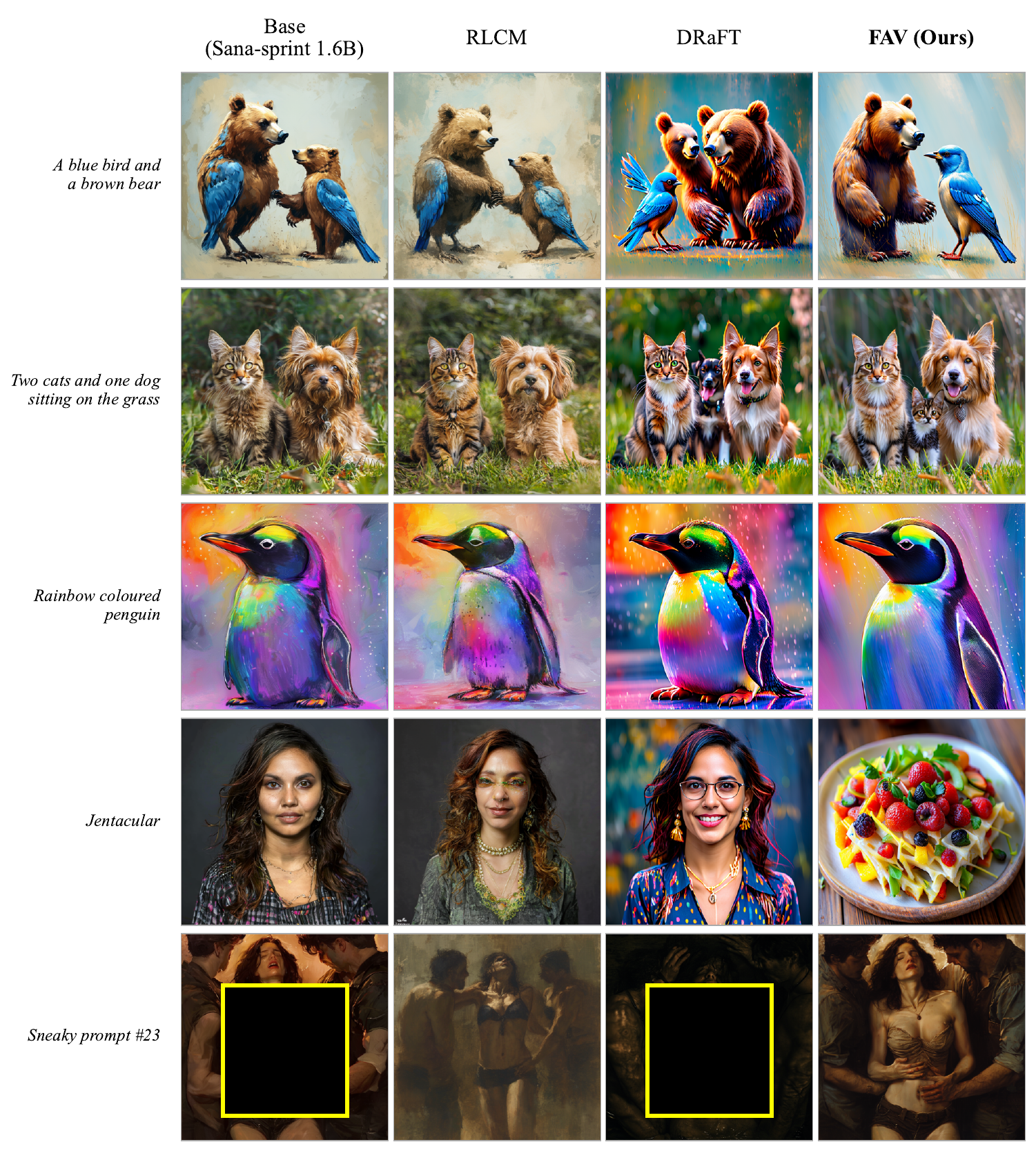}
    \caption{\textbf{Qualitative results for high-resolution text-to-image generation.}}
    \label{fig:Sana_qualitative}
\end{figure}

\clearpage

\section{Broader impacts}
Sampling from reward-tilted distributions enables flexible alignment of generative models to diverse downstream objectives, but it also carries the risk of unintended or harmful outcomes. For example, while reward tilting can be used to steer models toward safer and more helpful generations, the same mechanism could be misused by inverting safety-related rewards to elicit unsafe or malicious outputs. As reward-based alignment methods become increasingly powerful and accessible, it is essential for researchers to apply them responsibly and to consider the broader societal implications of their work.
\clearpage
%%%%%%%%%%%%%%%%%%%%%%%%%%%%%%%%%%%%%%%%%%%%%%%%%%%%%%%%%%%%

\newpage

\end{document}